\documentclass[journal]{IEEEtran}
\usepackage{cite}  
\usepackage{graphicx}
\usepackage{amsmath} 
\usepackage{amssymb}  
\usepackage{float}  
\usepackage{subcaption} 

\usepackage{multirow}
\usepackage{multicol}
\usepackage{booktabs}
\usepackage{tabularx}

\usepackage{xcolor}
\usepackage{hyperref}

\begin{document}
%
\title{SelfVIO: Self-Supervised Deep Monocular Visual-Inertial Odometry and Depth Estimation}
%
%
%

\author{Yasin Almalioglu$^{1}$, Mehmet Turan$^{2}$, Alp Eren Sar{\i}$^{3}$, Muhamad Risqi U. Saputra$^{1}$, Pedro P. B. de Gusmão$^{1}$, Andrew Markham$^{1}$, and Niki Trigoni$^{1}$
\thanks{$^{1}$Yasin Almalioglu, Muhamad Risqi U. Saputra, Pedro P. B. de Gusmão, Andrew Markham, and Niki Trigoni are with the Computer Science Department, The University of Oxford, UK
        {\tt\small \{yasin.almalioglu, muhamad.saputra, pedro.gusmao, andrew.markham, niki.trigoni\}@cs.ox.ac.uk}}%
\thanks{$^{2}$Mehmet Turan is with the Institute of Biomedical Engineering, Bogazici University, Turkey
        {\tt\small mehmet.turan@boun.edu.tr}}%
\thanks{$^{3}$Alp Eren Sar{\i} is with Department of Electrical and Electronics Engineering, Middle East Technical University, Ankara, Turkey
        {\tt\small asari@metu.edu.tr}}%
}

%



\maketitle

\begin{abstract}
In the last decade, numerous supervised deep learning approaches requiring large amounts of labeled data have been proposed for visual-inertial odometry (VIO) and depth map estimation.
To overcome the data limitation, self-supervised
learning has emerged as a promising alternative, exploiting constraints such as geometric and photometric consistency in the scene. 
In this study, we introduce a novel self-supervised deep learning-based VIO and depth map recovery approach (SelfVIO) using adversarial training and self-adaptive visual-inertial sensor fusion. 
SelfVIO learns to jointly estimate 6 degrees-of-freedom (6-DoF) ego-motion and a depth map of the scene from unlabeled monocular RGB image sequences and inertial measurement unit (IMU) readings. 
The proposed approach is able to perform VIO without the need for IMU intrinsic parameters and/or the extrinsic calibration between the IMU and the camera. 
We provide comprehensive quantitative and qualitative evaluations of the proposed framework comparing its performance with state-of-the-art VIO, VO, and visual simultaneous
localization and mapping (VSLAM) approaches on the KITTI, EuRoC and Cityscapes datasets.
Detailed comparisons prove that 
SelfVIO outperforms state-of-the-art VIO approaches in terms of pose estimation and depth recovery, making it a promising approach among existing methods in the literature.
\end{abstract}


%
\IEEEpeerreviewmaketitle

\section{Introduction}
\label{sec:intro}
\IEEEPARstart{E}{stimation} of ego-motion and scene geometry is
one of the key challenges in many engineering fields such as robotics and autonomous driving. 
In the last few decades, visual odometry (VO)  systems have attracted a substantial amount of attention due to low-cost hardware setups and rich visual representation \cite{fraundorfer2012visual}. 
{\color{black}
However, monocular VO is confronted with numerous challenges such as scale ambiguity, the need for hand-crafted mathematical features (e.g., ORB, BRISK), strict parameter tuning and image blur caused by abrupt camera motion, which might corrupt VO algorithms if deployed in low-textured areas and variable ambient lighting conditions \cite{engel2014lsd, mur2017orb}. 
}
For such cases, visual inertial odometry (VIO) systems increase the robustness of VO systems, incorporating
information from an inertial measurement unit (IMU)
to improve motion tracking performance \cite{mur2017visual, qin2018vins}.

\begin{figure}
\centering
\includegraphics[width=\columnwidth]{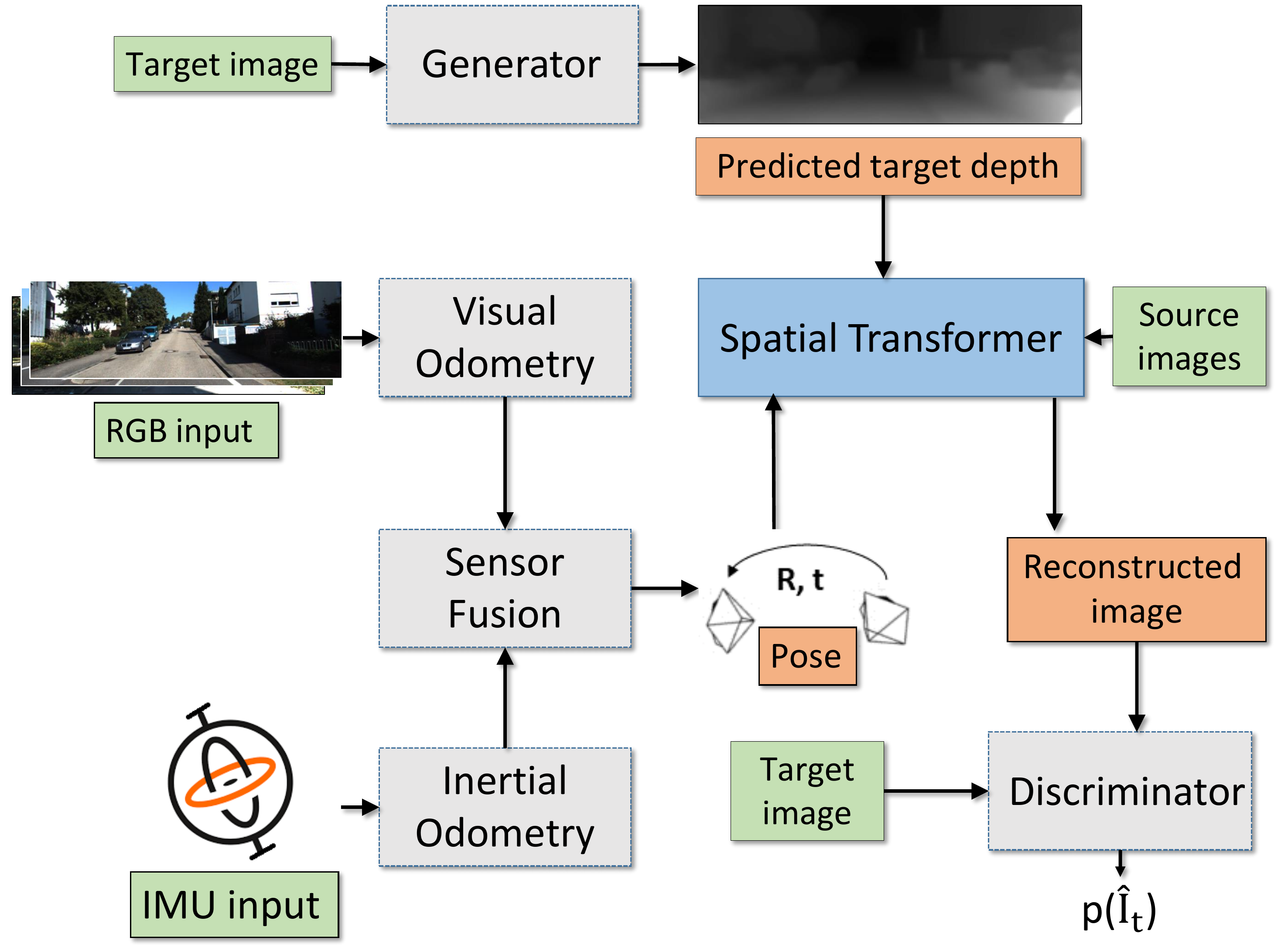}
\caption{\textbf{Architecture overview.} The proposed unsupervised deep learning approach consists of depth generation, visual odometry, inertial odometry, visual-inertial fusion, spatial transformer, and target discrimination modules. Unlabeled image sequences and raw IMU measurements are provided as inputs to the network. The method estimates relative translation and rotation between consecutive frames parametrized as 6-DoF motion and a depth image as a disparity map for a given view. The green and orange boxes represent inputs and intermediate outputs of the system, respectively.} 
\label{fig:model_overview}
\end{figure}  

Supervised deep learning methods have achieved state-of-the-art results on various computer vision problems using
large amounts of labeled data \cite{he2017mask,krizhevsky2012imagenet,long2015fully}.
Moreover, supervised deep VIO and depth recovery techniques have shown promising performance in challenging environments and successfully alleviate issues such as scale drift, need for feature extraction and parameter fine-tuning \cite{wang2017deepvo, clark2017vinet, turan2018deep, turan2019learning}. 
{\color{black}
Although learning based methods use raw input data similar to the dense VO and VIO methods, they also extract features related to odometry, depth and optical flow without explicit mathematical modeling \cite{engel2014lsd, mur2015orb, mur2017orb, clark2017vinet}.
}
Most existing deep learning approaches in the literature treat VIO and depth recovery as a
supervised learning problem, where they have color input
images, corresponding target depth values and relative transformation of images at training time. 
VIO as a regression problem in supervised deep learning exploits the capability of convolutional neural networks (CNNs) and recurrent neural networks (RNNs) to estimate camera motion, calculate optical flow, and  extract efficient feature representations from raw RGB and IMU input \cite{wang2017deepvo, clark2017vinet, muller2017flowdometry, turan2018deep}.
However, for many vision-aided localization and navigation
problems requiring dense, continuous-valued outputs (e.g. visual-inertial odometry (VIO) and depth map reconstruction), it is either impractical or expensive to acquire ground truth data for a large variety of scenes \cite{geiger2012we}.
Firstly, a state estimator uses timestamps for each camera image and IMU sample to enable the processing of the sensor measurements, which are typically taken either from the sensor itself, or from the operating system of the computer receiving the data.
{\color{black} 
However, a delay (different for each sensor) exists between the actual sampling of a measurement and its timestamp due to the time needed for data transfer, sensor latency, and OS overhead. Furthermore, even if hardware time synchronization is used for timestamping (e.g., different clocks on sensors), these clocks may suffer from clock skew, resulting in an unknown time offset that typically exists between the timestamps of the camera and the IMU \cite{qin2018online}.
}
Secondly, even when ground truth depth data
is available, it can be imperfect and cause distinct prediction artifacts. 
For example, systems employing rotating LIDAR scanners suffer from the need for tight temporal alignment between laser scans and corresponding camera images even if the camera and LIDAR are carefully synchronized \cite{asvadi2018multimodal}.
In addition, structured light depth sensors — and to
a lesser extent, LIDAR and time-of-flight sensors — suffer
from noise and structural artifacts, especially in the presence
of reflective, transparent, or dark surfaces. Last, there is
usually an offset between the depth sensor and the camera,
which causes shifts in the point cloud projection onto the camera viewpoint. 
These problems may lead to degraded performance and even failure for learning-based models trained on such data \cite{mahjourian2018unsupervised, turan2018endo}.

In recent years, unsupervised deep learning approaches have emerged to address the problem
of limited training data \cite{artetxe2017unsupervised, jason2016back, meister2018unflow}.
As an alternative, these approaches instead treat
depth estimation as an image reconstruction problem during
training. The intuition here is that, given a sequence of monocular images, we can learn a function that is able to
reconstruct a target image from source images, exploiting the 3D geometry of the scene.
To learn a mapping from pixels to depth and camera
motion without the ground truth is challenging because each of
these problems is highly ambiguous. To address this issue, recent studies imposed additional constraints and exploited the geometric relations
between ego-motion and the depth map \cite{zhou2017unsupervised, mahjourian2018unsupervised}.
Recently, optical flow has been widely studied and
used as a self-supervisory signal for learning an unsupervised ego-motion system, but it has an aperture problem due to the missing
structure in local parts of the single camera \cite{fortun2015optical}.
However,
most unsupervised methods learn only from photometric and temporal consistency between consecutive frames in monocular
videos, which are prone to overly smoothed depth map estimations.

To overcome these limitations, we propose a self-supervised VIO and depth map reconstruction system based on adversarial training and attentive sensor fusion (see Fig. \ref{fig:model_overview}), extending our GANVO work \cite{almalioglu2019ganvo}.
{\color{black}
GANVO is a generative unsupervised learning framework that predicts 6-DoF pose camera motion and a monocular depth map of the scene from unlabelled RGB image sequences, using deep convolutional Generative Adversarial Networks (GANs). 
Instead of ground truth pose and depth values, GANVO creates a supervisory signal by warping view sequences and assigning the re-projection minimization to the objective loss function that is adopted in multi-view pose estimation and single-view depth generation network.
In this work, we introduce a novel sensor fusion technique to incorporate motion information captured by an interoceptive and mostly environment-agnostic raw inertial data into loosely synchronized visual data captured by an exteroceptive RGB camera sensor.
}
Furthermore, we conduct experiments on the publicly available EuRoC MAV dataset \cite{burri2016euroc} to measure the robustness of the fusion system against miscalibration.
Additionally, we separate the
effects of the VO module from the pose estimates extracted from
IMU measurements to test the effectiveness of each module.
Moreover, we perform ablation studies  to compare the performance of convolutional and recurrent networks.
In addition to the results presented in \cite{almalioglu2019ganvo}, here we thoroughly evaluate the benefit of the adversarial generative approach.
In summary, the main contributions of the approach are as follows: 
\begin{itemize}
	\item To the best of our knowledge, this is the first self-supervised deep joint monocular VIO and depth reconstruction method in the literature; 	
	\item We propose a novel unsupervised sensor fusion technique for the camera and the IMU, which extracts and fuses motion features from raw IMU measurements and RGB camera images using convolutional and recurrent modules based on an attention mechanism;
	\item No strict temporal or spatial calibration between camera and IMU is necessary for pose and depth estimation, contrary to traditional VO approaches.
\end{itemize}

Evaluations made on the KITTI \cite{geiger2013vision}, EuRoC \cite{burri2016euroc} and Cityscapes \cite{cordts2016cityscapes} datasets prove the effectiveness of SelfVIO. The organization of this paper is as follows. Previous work in this domain is discussed in Section \ref{sec:rel_work}. Section \ref{sec:architecture} describes the proposed unsupervised deep learning architecture and its mathematical background in detail. Section \ref{sec:exp_setup} describes the experimental setup and evaluation methods. Section \ref{sec:res_discussion} shows and discusses detailed quantitative and qualitative results with comprehensive comparisons to existing methods in the literature. Finally, Section \ref{sec:conclusion} concludes the study with some interesting future directions.

\section{Related Work}
\label{sec:rel_work}
In this section, we briefly outline the related works focused on VIO including traditional and learning-based methods.
\subsection{Traditional Methods}
Traditional VIO solutions combine visual and inertial data in a single pose estimator and lead to more robust and higher accuracy compared to VO 
even in complex and dynamic environments. 
The fusion of camera images and
IMU measurements is typically accomplished by filter-based or optimization-based approaches.
{\color{black}
Early works of filter-based approaches formulated visual-inertial fusion as a pure sensor fusion problem, which fuses vision as an independent 6-DoF sensor with inertial measurements in a filtering framework (called loosely-coupled) \cite{weiss2012real}.
In a recent loosely-coupled method, Omari et al. proposed a filter-based direct stereo visual-inertial system, which fuses IMU with respect to the last keyframe.
These loosely coupled approaches allow modular integration of visual odometry methods without modification. 
However, more recent works follow
a tightly coupled approach to optimally exploit both sensor modalities, treating visual-inertial odometry
as one integrated estimation problem.
}
The multi-state constraint Kalman filter (MSCKF) \cite{mourikis2007multi} is a standard for filtering-based VIO approaches. It has low computational complexity that is linear
in the number of features used for ego-motion estimation. While MSCKF-based approaches are generally more robust compared to optimization-based approaches especially in large-scale real environments, they suffer from lower accuracy in comparison (as has been recently reported in \cite{delmerico2018benchmark}).
{\color{black}
Li et al. \cite{qin2018online} rigorously addressed online calibration for the first time based on MSCKF, unlike offline sensor to sensor spatial transformation and time offset calibration systems such as \cite{furgale2013unified}.
This online calibration method shows explicitly that the time offset is, in general, observable and provides the sufficient theoretical conditions for the observability of time offset alone, while practical degenerate motions are not thoroughly examined \cite{yang2019degenerate}.
Li et al. \cite{li2013high} proved that the standard method of computing Jacobian matrices in filters inevitably causes inconsistencies and accuracy loss. For example, they showed that the yaw errors of the MSCKF lay outside the $3\sigma$ bounds, which indicates filter inconsistencies.
They modified the MSCKF algorithm to ensure the correct observability properties without incurring additional computational costs.
}
ROVIO \cite{bloesch2017iterated}
is another filtering-based VIO algorithm for monocular cameras that utilizes the intensity errors in the update step of an extended Kalman filter (EKF) to fuse visual and inertial data. 
ROVIO uses a robocentric approach that estimates
3D landmark positions relative to the current
camera pose.

{\color{black}
On the other hand, optimization-based approaches operate based on an energy-function representation in a non-linear optimization framework. While the complementary nature of filter-based and optimization-based approaches has long been investigated \cite{eustice2006exactly}, energy-based representations \cite{jones2011visual, usenko2016direct} allows easy and
adaptive re-linearization of energy terms, which
avoids systematic error integration caused by linearization.
}
OKVIS \cite{leutenegger2015keyframe} is a widely used, optimization-based 
visual-inertial SLAM approach for monocular and stereo cameras.
OKVIS uses a nonlinear batch optimization on saved keyframes consisting of an image and
estimated camera pose. It updates a local map of landmarks to estimate camera motion without any loop closure constraint.
{\color{black}
To avoid repeated constraints caused by the parameterization of relative motion integration, Lupton et al. \cite{lupton2011visual} proposed IMU pre-integration to reduce computation, changing the IMU data between two frames by pre-integrating the motion constraints.
Forster et al. \cite{forster2015imu} further improve this principle by applying it to the visual-inertial SLAM framework to reduce bias.
Besides, systems that fused IMU data into the classic visual odometry also attracted widespread attention. Usenko et al. \cite{usenko2016direct} proposed a stereo direct VIO to combine IMU with stereo LSD-SLAM \cite{engel2015large}. They recovered the full state containing camera pose, translational velocity, and IMU biases of all frames, using a joint optimization method. Concha et al. \cite{concha2016visual} devised the first direct real-time tightly-coupled VIO algorithm, but the initialization was not introduced.
}
VINS-Mono \cite{qin2018vins} is a tightly coupled,
nonlinear optimization-based method for monocular cameras. It uses a pose graph optimization to enforce global consistency, which is constrained by a loop detection module. 
{\color{black}
VINS-Mono features efficient IMU pre-integration with bias correction, automatic initialization of estimator, online extrinsic calibration, failure detection, and loop detection.
}

\subsection{Learning-Based Methods}

{\color{black}
Eigen et al. \cite{eigen2014depth} proposed a two-scale deep network and showed that it was possible to produce dense pixel depth estimates, training on images, and the corresponding ground truth depth values. Unlike
most other previous work in single view depth estimation, their model learns a representation directly from the raw pixel values, without any need for handcrafted features or an initial over-segmentation. Several works followed the success of this approach using techniques such as the conditional random fields to improve the reconstruction accuracy \cite{li2015depth}, incorporating strong scene priors for surface normal estimation \cite{wang2015designing}, and the use of more robust loss functions \cite{laina2016deeper}. 
Again, like the most previous stereo methods, these approaches rely on existing high quality, pixel aligned, and dense ground truth depth maps at training time.

In recent years, several works adapt the classical GAN to estimate the depth from a single image \cite{cs2018monocular, aleotti2018generative, wu2019spatial}, following the success of
GANs in many learning based applications such as style transfer \cite{johnson2016perceptual}, image-to-image translation \cite{isola2017image}, image editing \cite{zhu2016generative} and cross-domain image generation \cite{bousmalis2017unsupervised}.
Pilzer et al. \cite{pilzer2018unsupervised} proposed a depth estimation model that employs the cycled generative networks to estimate depth from stereo pair in an unsupervised manner. These works demonstrate the effectiveness of GANs in-depth map estimation.
}

VINet \cite{clark2017vinet} was the first end-to-end trainable visual-inertial deep network. 
However, VINet was trained in
a supervised manner and thus required the ground truth pose
differences for each exemplar in the training set.
{\color{black}
Recently, there have been several successful unsupervised depth estimation approaches, which use image warping as part of reconstruction loss to create a supervision signal similar to our network.
} Garg et al. \cite{garg2016unsupervised}, Godard et al. \cite{godard2017unsupervised} and Zhan et al. \cite{zhan2018unsupervised} used such methods with stereo image pairs with known
camera baselines and reconstruction loss for training. Thus, while
technically unsupervised, stereo baseline effectively provides
a known transform between two images.

\begin{figure*}
\centering
\includegraphics[width=\textwidth]{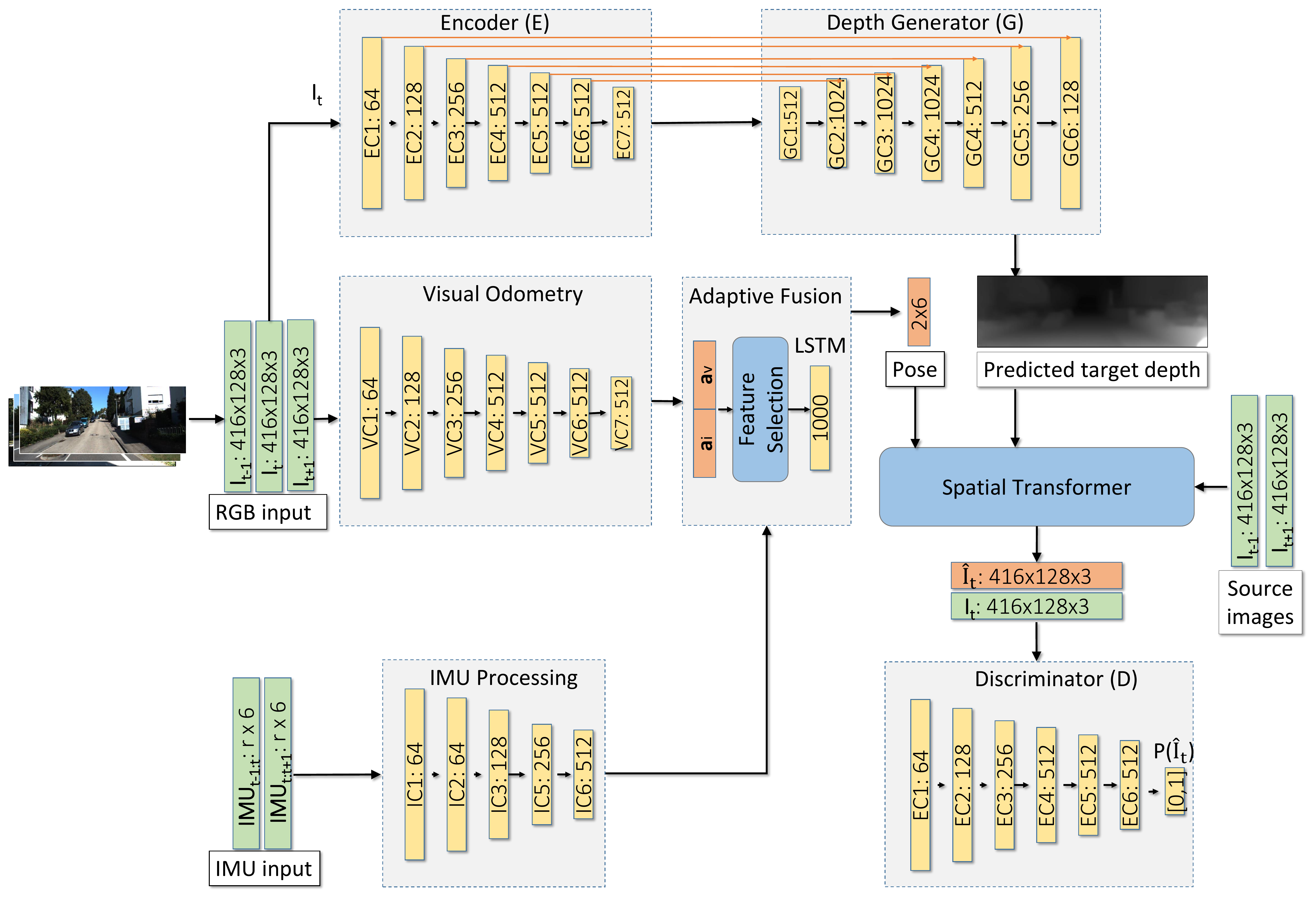}
\caption{\textbf{The proposed architecture for pose estimation and depth map generation}. The spatial dimensions of layers and output channels are proportional to the tensor shapes that flow through the network. Generator network $G$ maps the feature vector generated by the encoder network $E$ to the depth image space. In parallel, the visual odometry module extracts VO-related features through a convolutional network, while the inertial odometry module estimates inertial features related to ego-motion. The adaptive sensor fusion module fuses visual and inertial information, and estimates pose using a recurrent network that captures temporal relations among the input sequences. Pose results are collected after adaptive fusion operation, which has $6*(N-1)$ output channels for 6-DoF motion parameters, where $N$ is the length of the input sequence. The spatial transformer module reconstructs the target view using the estimated depth map and pose values. The discriminator $D$ maps the reconstructed RGB image to a likelihood of the target image, which determines whether it is the reconstructed or original target image.}
\label{fig:deeparchitecture}
\end{figure*}

More recent works \cite{mahjourian2018unsupervised, ummenhofer2017demon, yin2018geonet, zhou2017unsupervised} have formulated odometry and depth estimation problems by coupling two or more problems together in an unsupervised learning framework.
Zhou et al. \cite{zhou2017unsupervised} introduced joint unsupervised learning of ego-motion and depth from multiple unlabeled RGB frames.
They input
a consecutive sequence of images and output a change in pose
between the middle image of the sequence and every other image
in the sequence, and the estimated depth of the middle image.
A recent work \cite{wulff2018temporal} used a more
explicit geometric loss to jointly learn depth and camera motion for rigid scenes with a
semi-differentiable iterative closest point (ICP) module.
These VO approaches estimate ego-motion only by the spatial information existing in several
frames, which means temporal information within the frames
is not fully utilized. As a result, the estimates are
inaccurate and discontinuous.

UnDeepVO \cite{li2018undeepvo} is another unsupervised depth
and ego-motion estimation work. 
{\color{black}
It differs from \cite{zhou2017unsupervised} in that it can estimate the camera trajectory on an absolute scale.
} However, unlike
\cite{zhou2017unsupervised} and similar to \cite{garg2016unsupervised, godard2017unsupervised}, it uses stereo image pairs for training
where the baseline between images is available and thus, UnDeepVO
can only be trained on datasets where stereo image pairs are
existent. 
{\color{black}
Additionally, stereo images are recorded simultaneously, and the spatial transformation between
paired images from stereo cameras are unobservable by an IMU. Thus, the network architecture of UnDeepVO
cannot be extended to include motion estimates derived from
inertial measurements.
}
VIOLearner \cite{shamwell2019unsupervised} is a recent unsupervised learning-based approach to
VIO using multiview RGB-depth (RGB-D) images, which extends the work of \cite{shamwell2018vision}. 
It uses a learned optimizer to minimize photometric
loss for ego-motion estimation, which leverages 
the Jacobians of scaled image projection
errors with respect to a spatial grid of pixel coordinates
similar to \cite{clark2018learning}.
Although no ground truth odometry data are needed, the depth input to the system provides external supervision to the network, which may not always be available.

One critical issue of these unsupervised works is
the fact that they use auto encoder-decoder-based traditional
depth estimators with a tendency to generate overly smooth
images \cite{dosovitskiy2016generating}.
{\color{black}
GANVO \cite{almalioglu2019ganvo} is the first unsupervised adversarial generative approach to jointly estimate multiview pose and monocular depth map. GANVO solves the smoothness problem in the reconstructed depth maps using GANs. Therefore, we apply GANs to provide sharper and more accurate depth maps, extending the work of \cite{almalioglu2019ganvo}.
} 
The second issue of
the aforementioned unsupervised techniques is the fact that
they solely employ CNNs that only analyze just-in-moment
information to estimate camera pose \cite{wang2017deepvo, turan2018deep, almalioglu2019milli}. We address
this issue by employing a CNN-RNN architecture to capture
temporal relations across frames.
Furthermore, these existing VIO works use a direct fusion approach that concatenates all features extracted from different modalities, resulting in sub-optimal performance, as not
all features are useful and necessary \cite{chen2019selective}. We introduce an attention mechanism to self-adaptively fuse the different modalities conditioned on the input data. We discuss our reason behind these design choices in the related sections.


\section{Self-supervised Monocular VIO and Depth Estimation Architecture}
\label{sec:architecture}

Given unlabeled monocular RGB image sequences and raw IMU measurements, the proposed approach learns a function $f$ that regresses 6-DoF camera motion and predicts the per-pixel scene depth. 
An overview of our SelfVIO architecture is depicted in Fig. \ref{fig:model_overview}.
{\color{black}
We stack the monocular RGB sequences consisting of a target view ($\mathbf{I}_t$) and source views ($<\mathbf{I}_{t-1}, \mathbf{I}_{t+1}>$) to form an input batch for the multiview visual odometry module.
} 
The VO module consisting of convolutional layers regresses the relative 6-DoF pose values of the source views with respect to the target view.
We form an IMU input tensor using raw linear acceleration and angular velocity values measured by an IMU between $t-1$ and $t+1$, which is processed in the inertial odometry module to estimate the relative motion of the source views.
We fuse the 6-DoF pose values estimated by visual and inertial odometry modules in a self-adaptive fusion module, attentively selecting certain features that are significant for pose regression.
{\color{black}
In parallel, the target view ($\mathbf{I}_t$) is fed into the encoder module.
} The depth generator module estimates a depth map of the target view  by inferring the disparities that warp the source views to the target. The spatial transformer module synthesizes the target image using the generated depth map and the nearby color pixels in a source image sampled at locations determined by a fused 3D affine transformation.
The geometric constraints that provide a supervision signal cause the neural network to synthesize a target image from multiple source images acquired from different camera poses. 
The view discriminator module learns to distinguish difference between a fake (synthesized by the spatial transformer) and a real target image.
In this way, each subnetwork targets a specific subtask and the complex scene geometry understanding goal is decomposed into smaller subgoals.  

In the overall adversarial paradigm, 
a generator network is trained to produce output that cannot be
distinguished from the original image by an adversarially optimized
discriminator network.
The objective of the generator is to trick the discriminator, i.e. to generate a depth map of the target view such that the discriminator cannot distinguish the reconstructed view from the original view. 
Unlike the typical use of GANs, the spatial transformer module maps the output image of the generator to the color space of the target view and the discriminator classifies this reconstructed colored view rather than the direct output of the generator.
The proposed scheme enables us to predict the relative motion and depth map in an unsupervised manner, which is explained in the following sections in detail.

\subsection{Depth Estimation}
The first part of the architecture is the depth generator network that synthesizes a single-view depth map by translating the target RGB frame.
A defining feature of image-to-depth translation problems
is that they map a high-resolution input tensor to a high resolution output tensor, which differs in surface appearance. However, both images
are renderings of the same underlying structure. Therefore,
the structure in the RGB frame is roughly aligned with the structure in the depth map. 

The depth generator network is based on a GAN design that learns the underlying generative model of the input image $p(\mathbf{I}_t)$. Three subnetworks are involved in the adversarial depth generation process: an encoder network $E$, a generator network $G$, and a discriminator network $D$. The encoder $E$ extracts a feature vector $\mathbf{z}$ from the input target image $\mathbf{I}_t$, i.e. $E(\mathbf{I}_t) = \mathbf{z}$. $G$ maps the vector $\mathbf{z}$ to the depth image space which is used in spatial transformer module to reconstruct the original target view. $D$ classifies the reconstructed view as synthesized or real.

Many previous solutions \cite{zhou2017unsupervised, li2018undeepvo, ranjan2019competitive} to the single-view depth estimation are based on an encoder-decoder network \cite{hinton2006reducing}. 
Such a network passes the input through a series of layers that progressively downsample until a bottleneck layer and, then,
the process is reversed by upsampling. 
All information flow passes through all the layers,
including the bottleneck. For the image-to-depth translation problem, there is a great deal of low-level information shared
between the input and output, and the network should make use of this information by directly sharing it across the layers. 
As an example, RGB image input and the depth map output share the location of prominent edges.
To enable the generator to circumvent the bottleneck for such shared low-level information, we add skip connections similar to the general shape of a U-Net \cite{ronneberger2015u}.
Specifically, these connections are placed between each layer $i$ and layer $n-i$, where $n$ is the total number of layers, which concatenate all channels at layer $i$ with those at layer $n-i$.

\subsection{Visual Odometry}
{\color{black}
The VO module (see Fig. \ref{fig:deeparchitecture}) is designed to take two concatenated source views and a target view along the color channels as input and to output a visual feature vector $\mathbf{p}_V$ introduced by motion and temporal dynamics across frames. 
The network is composed of 7 stride-2 convolutions followed by the adaptive fusion module. 
We decouple the convolution layer for translation and
rotation using the shared weights as it has been shown to work better in separate branches as in
\cite{saputra2019deeptio}. We also use a dropout \cite{srivastava2014dropout} between the convolution
layers at the rate of $0.25$ to help regularization.
The last convolution layer gives a visual feature vector to encode geometrically meaningful features for movement estimation, which is used to define the 3D affine transformation between target image $\mathbf{I}_t$ and source images $\mathbf{I}_{t-1}$ and $\mathbf{I}_{t+1}$. 
}

\subsection{Inertial Odometry}
\label{sec:inertial_odometry}
SelfVIO takes raw IMU measurements in the following form:
\[
\mathbf{M}=
\begin{bmatrix}
    \mathbf{\alpha}_{t-1}       & \mathbf{\omega}_{t-1}  \\
    \dots       & \dots  \\
    \mathbf{\alpha}_{t+1}       & \mathbf{\omega}_{t+1} 
\end{bmatrix}
\in \mathbb{R}^{n\times6},
\]
where $\mathbf{\alpha}\in \mathbb{R}^{3}$ is linear acceleration, $\mathbf{\omega}\in \mathbb{R}^{3}$ is angular velocity, and $n$ is the number of IMU samples obtained between time $t-1$ and $t+1$ 
{\color{black}
(no timestamp related to the IMU of the camera is passed to the network).
The IMU module receives the same size of padded input in each time frame.
}
The IMU processing module of SelfVIO uses two parallel branches consisting of
$5$ convolutional layers for the IMU angular velocity and linear
acceleration (see Fig. \ref{fig:deeparchitecture} for more detail). 
Each branch on the IMU measurements has the following convolutional layers:
\begin{enumerate}
\item two layers: $64$ single-stride filters with kernel size $3\times5$,
\item one layer: $128$ filters of stride $2$ with kernel size $3\times5$,
\item one layer: $256$ filters of stride $2$ with kernel size $3\times5$, and
\item one layer: $512$ filters of stride $2$ with kernel size $3\times2$.
\end{enumerate}
The outputs of the final angular velocity and linear
acceleration branches were flattened into $2\times3$ tensors using
a convolutional layer with three filters of kernel size $1$ and stride
$1$ before they are concatenated into a tensor $\mathbf{p}_M$. Thus, it learns to estimate 3D affine transformation between times $t-1$ and $t+1$.

\subsection{Self-Adaptive Visual-Inertial Fusion}
In learning-based VIO, a standard method for fusion is concatenation of feature vectors coming from different modalities, which may result in suboptimal performance, as not
all features are equally reliable \cite{chen2019selective}. 
For example, the fusion is plagued by the intrinsic noise distribution of each modality such as white random noise and sensor bias in IMU data. 
Moreover, many real-world applications suffer from
poor calibration and synchronization between different modalities.
To eliminate the effects of these factors, we employ an attention mechanism \cite{vaswani2017attention}, which allow the network to automatically learn the best suitable feature combination given visual-inertial feature inputs.

The convolutional layers of the VO and IMU processing modules extract features from the input sequences and estimate ego-motion, which is propagated to the self-adaptive fusion module.
In our attention mechanism, we use a deterministic soft
fusion approach to attentively fuse features. 
{\color{black}
The adaptive fusion module learns
visual ($\mathbf{s}_v$) and inertial ($\mathbf{s}_i$) filters to reweight each feature by conditioning on all channels:
\begin{equation}
\mathbf{s}_v = \sigma(\mathbf{W}_v[\mathbf{a}_v,\mathbf{a}_i])
\end{equation}
\begin{equation}
\mathbf{s}_i = \sigma(\mathbf{W}_i[\mathbf{a}_v,\mathbf{a}_i]),
\end{equation}
where $\sigma(x)=1/(1+e^{-x})$ is the sigmoid function, $[\mathbf{a}_v,\mathbf{a}_i]$ is the concatenation of all channel features, and $\mathbf{W}_v$ and $\mathbf{W}_i$ are the weights for each modality.
}
We multiply the visual and inertial features with these masks to weight the relative importance of the features:
\begin{equation}
\mathbf{W}_{fused} = [\mathbf{a}_v\odot \mathbf{s}_v, \mathbf{a}_i\odot \mathbf{s}_v],
\end{equation}
where $\odot$ is the elementwise multiplication. The resulting weight matrix $\mathbf{W}_{fused}$ is fed into
the RNN part (a two-layer bi-directional LSTM). 
The LSTM takes the combined feature
representation and its previous hidden states as input, and models the dynamics and connections between a
sequence of features.
After the recurrent network, a fully connected layer regresses the fused pose, which maps the
features to a 6-DoF pose vector.
It outputs $6*(N-1)$ ($N$ is the number of input views, i.e. $3$ ) channels for 6-DoF pose values for translation and rotation parameters, representing the motion over a time window $t-1$ and $t+1$. 
{\color{black}
The output pose vector defines the 3D affine transformation between target image $\mathbf{I}_t$ and source images $\mathbf{I}_{t-1}$ and $\mathbf{I}_{t+1}$. The LSTM improves the sequential learning capacity of the network, resulting in more accurate pose estimation.
}

\subsection{Spatial Transformer}

A sequence of $3$ consecutive frames is given to the pose network as input. An input sequence is denoted by $<I_{t-1}, I_{t}, I_{t+1}>$ where $t>0$ is the time index, $I_{t}$ is the target view, and the other frames are source views $I_{s} = <I_{t-1}, I_{t+1}>$ that are used to render the target image according to the objective function:
\begin{equation}
\mathcal{L}_{g} = \sum_{s}\sum_p | I_t (p) - \hat{I}_s(p) |
\end{equation}
where $p$ is the pixel coordinate index, and $\hat{I}_s$ is the projected image of the source view $I_s$ onto the target coordinate frame using a depth image-based rendering module. 
For the rendering, we define the static scene geometry by a collection
of depth maps $D_i$ for frame $i$ and the relative camera motion $T_{t\rightarrow s}$ from the target to the source frame. The relative 2D rigid
flow from target image $I_t$ to source image $I_s$ can be represented by\footnote{Similar to~\cite{zhou2017unsupervised}, we omit the necessary conversion to homogeneous coordinates for notation brevity.}:
\begin{equation}
    \label{equa::proj}
f_{t\to s}^{rig}(p_t) = KT_{t\to s}D_t(p_t)K^{-1}p_t - p_t,
\end{equation}
where $K$ denotes the $4\times 4$ camera transformation matrix and $p_t$ denotes homogeneous coordinates of pixels in target frame $I_t$.

We interpolate the nondiscrete $p_s$ values to find the expected intensity value at that position, using bilinear interpolation with the $4$ discrete neighbors of $p_s$ \cite{zhou2016view}. The mean intensity value for projected pixel is estimated as follows:
\begin{equation}
\hat{I}_s(p_t) = I_s(p_s) = \sum_{i\in\{top,bottom\}, j\in\{left,right\}}w^{ij}I_s(p_s^{ij})
\end{equation}
where $w^{ij}$ is the proximity value between the projected and neighboring pixels, which sums up to $1$. 
Guided by these positional constraints, we can apply differentiable inverse warping \cite{jaderberg2015spatial} between nearby frames, which later becomes the foundation of our self-supervised learning scheme.

\begin{figure*}
\centering
\includegraphics[width=\textwidth]{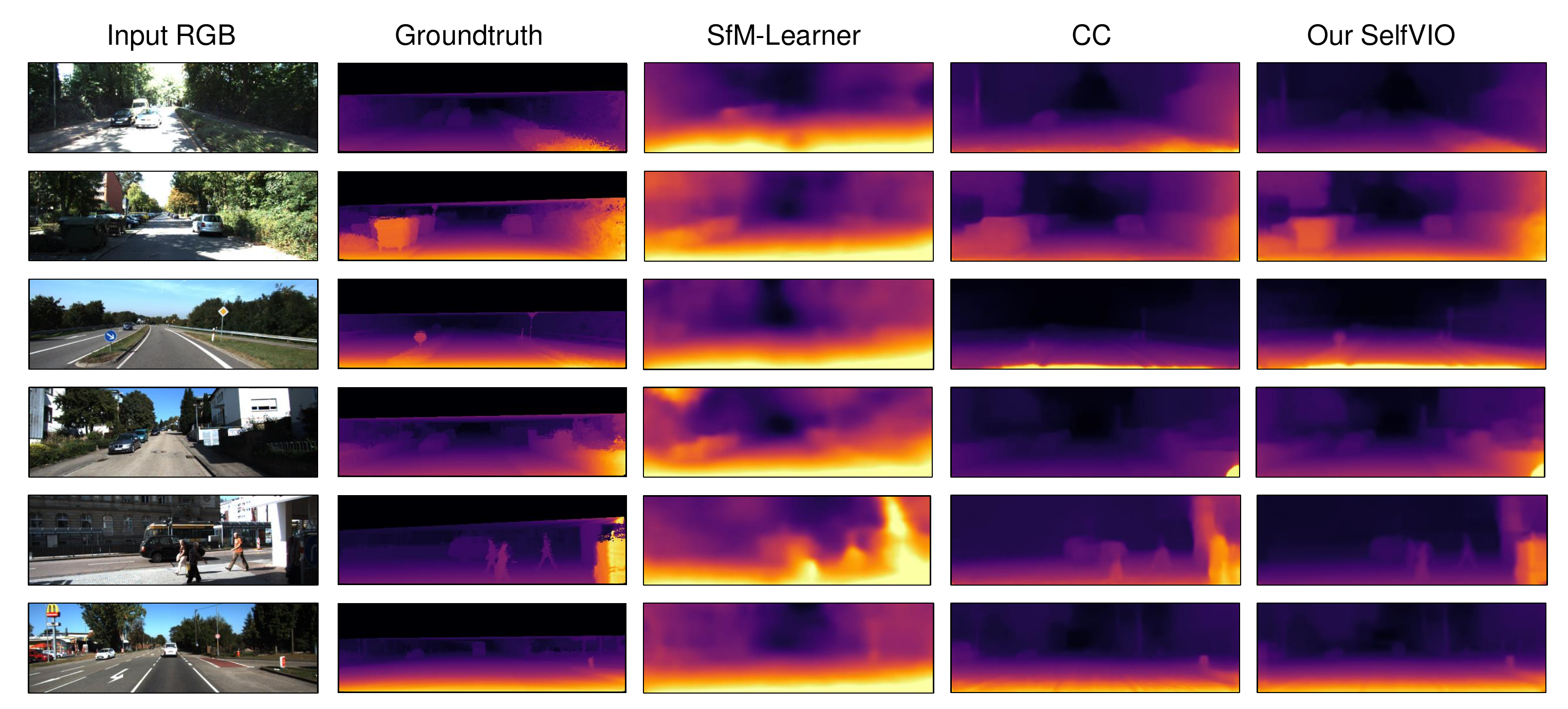}
\caption{\textbf{Qualitative results for monocular depth map estimation.} Comparison of unsupervised monocular depth estimation between SfM-Learner \cite{zhou2017unsupervised}, CC \cite{ranjan2019competitive} and the proposed SelfVIO. To visualize the ground truth depth map, we interpolated the sparse LIDAR point clouds and projected them onto the camera imaging
plane using the provided KITTI extrinsic and camera calibration
matrices. As seen in the figure, SelfVIO captures details in challenging scenes containing low-textured areas, shaded regions, and uneven road lines, preserving sharp, accurate and detailed depth map predictions both in close and distant regions.}
\label{fig:res_depth}
\end{figure*}

\begin{figure*}
\centering
\includegraphics[width=\textwidth]{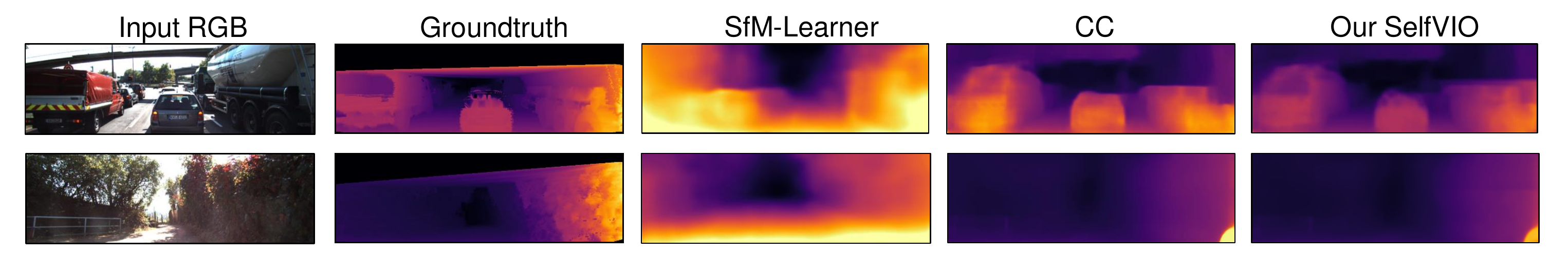}
\caption{\textbf{Degradation in depth reconstruction.} The performance of the compared methods SfM-Learner \cite{zhou2017unsupervised}, CC \cite{ranjan2019competitive} and the proposed SelfVIO degrades under challenging conditions such as vast open rural scenes and huge objects occluding the camera view.}
\label{fig:res_depth_degredation}
\end{figure*}

\subsection{View Discriminator}
The L2 and L1 losses produce blurry results on image generation problems \cite{larsen2016autoencoding}. Although these losses fail to encourage high-frequency crispness, in many cases they nonetheless accurately capture the low frequencies.
This motivates restricting the GAN discriminator to
model high-frequency structure, relying on an L1 term to
force low-frequency correctness. To model
high frequencies, it is sufficient to restrict our attention to
the structure in local image patches. Therefore, we employ
the PatchGAN \cite{isola2017image} discriminator architecture that only penalizes the structure at the scale of patches. This
discriminator tries to classify each $M\times M$ patch in an image as real or fake. We run this discriminator convolutionally across the image, averaging all responses to provide the
ultimate output of $D$.
Such a discriminator effectively models the image as a
Markov random field, assuming independence between pixels separated by more than a patch diameter. This connection was previously explored in  \cite{li2016precomputed}, and is also the common assumption in models of texture \cite{gatys2015texture}, which can be interpreted as a form of texture loss.

The spatial transformer module synthesizes a realistic image by the view reconstruction algorithm using the depth image generated by $G$ and estimated pose value.
$D$ classifies the input images sampled from the target data distribution $p_{data}$ into the fake and real categories, playing an adversarial role. These networks are trained by optimizing the objective loss function:
\begin{equation}
\begin{split}
\mathcal{L}_{d} = \min_{G} \max_{D} V(G,D) = &\mathbb{E}_{\textbf{I} \sim p_{data}(\textbf{I})} [\log(D(\textbf{I}))] + \\
&\mathbb{E}_{\textbf{z} \sim p(\textbf{z})} [\log(1-D(G(\textbf{z})))],
\end{split}
\end{equation}
where $\mathbf{I}$ is the sample image from the $p_{data}$ distribution and $\mathbf{z}$ is a feature encoding of $\mathbf{I}$ on the latent space. 

\subsection{The Adversarial Training}
In contrast to the original GAN \cite{goodfellow2014generative}, we remove fully connected hidden layers for deeper architectures and use batchnorms in the $G$ and $D$ networks. 
We replace pooling layers with strided convolutions and fractional-strided convolutions in $D$ and $G$ networks, respectively.
For all layer activations, we use LeakyReLU and ReLU  in the $D$ and $G$ networks, respectively, except for the output layer that uses tanh nonlinearity.
The GAN with these modifications and loss functions generates nonblurry depth maps and resolves the convergence problem during the training \cite{radford2015unsupervised}. The final objective for the optimization during the training is:
\begin{equation}
\mathcal{L}_{final} = \mathcal{L}_{g} + \beta \mathcal{L}_{d}
\end{equation}
where $\beta$ is the balance factor that is experimentally found to be optimal by the ratio between the expected values $\mathcal{L}_{g}$ and $\mathcal{L}_{d}$ at the end of the training.


\section{Experimental Setup}
\label{sec:exp_setup}

In this section, the datasets used in the experiments, network training protocol, evaluation methods are introduced including ablation studies, and performance evaluation in cases of poor intersensor calibration.

\begin{table*}
\begin{center}
\caption{Results on Depth Estimation. Supervised methods are shown in the first three rows. Data refers to the training set: Cityscapes (cs) and KITTI (k). For the experiments involving CS dataset, SelfVIO is trained without IMU as CS dataset lacks IMU data. }
\begin{tabular}{lccccc|ccc}
\toprule
& & \multicolumn{4}{c|}{Error (m)} & \multicolumn{3}{c}{Accuracy, $\delta$ } \\ \cline{3-6}  \cline{7-9}
Method& Data &AbsRel&SqRel&RMS&RMSlog &$<1.25$&$<1.25^2$& $<1.25^3$\\ 
\midrule
{ Eigen et al.~\cite{eigen2014depth} coarse}& k & 0.214& 1.605& 6.563& 0.292& 0.673& 0.884 &0.957\\
{ Eigen et al.~\cite{eigen2014depth} fine}& k & 0.203 &1.548& 6.307& 0.282& 0.702& 0.890& 0.958 \\
{ Liu et al. \cite{liu2016learning}} & k & 0.202& 1.614& 6.523& 0.275& 0.678& 0.895& 0.965\\
\midrule
{ SfM-Learner~\cite{zhou2017unsupervised}} &cs+k &0.198 & 1.836 & 6.565 & 0.275 & 0.718 & 0.901 & 0.960 \\
{ Mahjourian et al.~\cite{mahjourian2018unsupervised} }&cs+k & 0.159 & 1.231 & 5.912 & 0.243 & 0.784 & 0.923 & 0.970 \\
{ Geonet \cite{yin2018geonet} }&cs+k &{0.153} & 1.328 & 5.737 & 0.232 & 0.802 & {0.934} & 0.972 \\
{ DF-Net \cite{zou2018df} }&cs+k & 0.146 & 1.182 & 5.215 & \textbf{0.213} & 0.818 & 0.943 & \textbf{0.978}\\
{ CC~\cite{ranjan2019competitive} } & cs+k & 0.139 & 1.032 & 5.199 & \textbf{0.213} & 0.827 & 0.943&  0.977 \\
{ GANVO~\cite{almalioglu2019ganvo} } & cs+k & \textbf{0.138} & 1.155 & 4.412 & 0.232 & 0.820 & 0.939 & 0.976\\
{ SelfVIO (ours, no-IMU)} & cs+k & \textbf{0.138} & \textbf{1.013} & \textbf{4.317} & 0.231 & \textbf{0.849} & \textbf{0.958}&  { 0.979} \\
\midrule
{ SfM-Learner~\cite{zhou2017unsupervised}} &k &0.183 & 1.595 & 6.709 & 0.270 & 0.734 & 0.902 & 0.959 \\
{ Mahjourian et al.~\cite{mahjourian2018unsupervised}}&k & 0.163 & 1.240 & 6.220 & 0.250 & 0.762 & 0.916 & 0.968 \\
{ Geonet~\cite{yin2018geonet} }&k &{0.155} & 1.296 & 5.857 & 0.233 & {0.793} & {0.931} & {0.973} \\
{ DF-Net~\cite{zou2018df} }&k & 0.150  &1.124&  5.507& 0.223& 0.806& 0.933& 0.973 \\
{ CC~\cite{ranjan2019competitive}} & k &  0.140&     1.070&     5.326&     \textbf{0.217}&     0.826&     0.941&   0.975 \\
{ GANVO~\cite{almalioglu2019ganvo}} & k &  0.150 & 1.141 & 5.448 & 0.216 & 0.808 & 0.939 & 0.975\\
{ SelfVIO (ours)} & k &  \textbf{0.127}&     \textbf{1.018}&     \textbf{5.159}&     0.226&     \textbf{0.844}&     \textbf{0.963}&    \textbf{0.984} \\
\bottomrule
\end{tabular}
\label{tab:depth}
\end{center}
\end{table*}  

\subsection{Datasets}
\subsubsection{KITTI} 
The KITTI odometry dataset \cite{geiger2013vision} is a benchmark for depth and odometry evaluations including vision and LIDAR-based approaches. Images are recorded at $10$ Hz via an onboard camera mounted on a Volkswagen Passat B6. Frames are recorded in various environments such as residential, road, and campus scenes adding up to a $39.2$ km travel length.
{\color{black} 
Ground truth pose values at each camera exposure are determined using an OXTS RT 3003 GPS solution with an accuracy of $10$ cm. The corresponding ground truth pixel depth values are acquired via a Velodyne laser scanner.
A temporal synchronization between sensors is provided using a software-based calibration approach, which causes issues for VIO approaches that require strict time synchronization between RGB frames and IMU data.
}

We evaluate SelfVIO on the KITTI odometry dataset using Eigen et al.'s split \cite{eigen2014depth}. 
We use sequences $00-08$ for training and $09-10$ for the test set that is consistent across
related works  \cite{eigen2014depth, liu2016learning, almalioglu2019ganvo, mahjourian2018unsupervised, yin2018geonet, zhou2017unsupervised, zou2018df}. 
Additionally, $5\%$ of
KITTI sequences $00-08$ are withheld as a validation set, which leaves
a total of $18,422$ training images, $2,791$ testing images, and
$969$ validation images.
Input images are scaled to $256\times 832$ for training, whereas they are not limited to any specific image size at test time.
In all experiments, we randomly select an image for the
target and use consecutive images for the source. Corresponding
$100$ Hz IMU data are collected from the KITTI raw datasets and
for each target image, the preceding $100$ ms and the following
$100$ ms of IMU data are combined yielding a tensor of size $20\times 6$ 
($100$ ms between the source images and target).
Thus, the network learns how to implicitly
estimate a temporal offset between camera and IMU as well
as to determine an estimate of the initial velocity at the time of target image timestamp by looking to corresponding IMU data.

\subsubsection{EuRoC}
The EuRoC dataset \cite{burri2016euroc} contains 11 sequences recorded onboard from
an AscTec Firefly micro aerial vehicle (MAV) while it was manually piloted around three different indoor
environments executing 6-DoF motions. Within each environment, the sequences increase qualitatively in difficulty with increasing sequence
numbers. For example, Machine Hall 01 is "easy", while
Machine Hall 05 is a more challenging sequence in the
same environment, containing faster and loopier motions,
poor illumination conditions etc.
{\color{black}
We evaluate SelfVIO on the EuRoC odometry dataset using MH02(E), MH04(D), V103(D), V202(D) for testing, and the remaining sequences for training. 
Additionally, $5\%$ of the training sequences are withheld as a validation set.
}
All the EuRoC sequences are recorded by a front-facing visual-inertial sensor unit with tight synchronization between the stereo camera and IMU timestamps  captured using a MAV. 
Accurate ground truth is provided by laser
or motion capture tracking depending on the sequence, which has been used in many of the existing partial
comparative evaluations of VIO methods.
The dataset provides
synchronized global shutter WVGA stereo images at a rate of $20$ Hz that we use only the left camera image, and
the acceleration and angular rate measurements captured by a Skybotix VI IMU sensor at $200$ Hz.
In the Vicon
Room sequences, ground truth positioning measurements are provided by Vicon motion capture systems, while in  the Machine
Hall sequences, ground truth is
provided by a Leica MS50 laser tracker. 
The dataset containing sequences, ground truth and
sensor calibration data is publicly available \footnote{http://projects.asl.ethz.ch/datasets/doku.php?id=kmavvisualinertialdatasets}.
The EuRoC dataset, being recorded indoors on unstructured paths, exhibits motion blur and the trajectories follow highly irregular paths unlike the KITTI dataset.

\subsubsection{Cityscapes}
The Cityscapes Urban Scene 2016 dataset \cite{cordts2016cityscapes} is a large-scale dataset  mainly used
for semantic urban scene understanding, which contains $22,973$ stereo images for autonomous
driving in an urban environment collected in street scenes from 50 different
cities across Germany spanning several months.
The dataset also provides precomputed disparity depth maps
associated with the RGB images. 
Although it has a similar setting to the KITTI dataset, 
the Cityscapes dataset has higher resolution ($2048\times 1024$), more image quality, and
variety. 
We cropped the input images to keep only the top 80\% of the image,
removing the very reflective car hoods.

\begin{table}
  \centering
  \caption{
{\color{black}
Monocular VO results with our proposed SelfVIO evaluated on the training sequences. No loop closure is performed in the methods listed in the table. 
Note that monocular VISO2 and ORB-SLAM (without loop closure) did not work with image resolution $416 \times 128$, the results were obtained with full image resolution $1242\times376$.
}
7-DoF (6-DoF + scale) alignment with the ground-truth is applied for SfMLearner and monocular ORB-SLAM.}
    \begin{tabularx}{\columnwidth}{lXXXXXl|X}
    \toprule
     &  & Seq.00 & Seq.02    & Seq.05 & Seq.07 & Seq.08 & Mean \\
    \hline
    \multirow{2}{*}{SelfVIO} & \multicolumn{1}{|c|}{$t_{\text{rel}} {(\%)}$}     & \textbf{1.24} & \textbf{0.80} & \textbf{0.89} & 0.91  & \textbf{1.09}  & \textbf{0.95} \\
          & \multicolumn{1}{|c|}{$r_{\text{rel}} (^{\circ})$}     & \textbf{0.45} & \textbf{0.25} & 0.63  & \textbf{0.49}  & \textbf{0.36}  & \textbf{0.44} \\
    \hline
    \multirow{2}{*}{VIOLearner} & \multicolumn{1}{|c|}{$t_{\text{rel}} {(\%)}$}      & 1.50  & 1.20 & 0.97  & \textbf{0.84}  & 1.56  & 1.21 \\
          & \multicolumn{1}{|c|}{$r_{\text{rel}} (^{\circ})$}     & 0.61  & 0.43 & \textbf{0.51} & 0.66  & 0.61  & 0.56 \\
    \hline
    \multirow{2}{*}{UnDeepVO} & \multicolumn{1}{|c|}{$t_{\text{rel}} {(\%)}$}      & 4.14  & 5.58 & 3.40  & 3.15  & 4.08  & 4.07 \\
          & \multicolumn{1}{|c|}{$r_{\text{rel}} (^{\circ})$}     & 1.92  & 2.44 & 1.50  & 2.48  & 1.79  & 2.03 \\
    \hline
    \multirow{2}{*}{SfMLearner} & \multicolumn{1}{|c|}{$t_{\text{rel}} {(\%)}$}      & 65.27 & 57.59 & 16.76 & 17.52 & 24.02 & 36.23 \\
          & \multicolumn{1}{|c|}{$r_{\text{rel}} (^{\circ})$}     & 6.23  & 4.09 & 4.06  & 5.38  & 3.05  & 4.56 \\
    \hline
    \multirow{2}{*}{VISO2} & \multicolumn{1}{|c|}{$t_{\text{rel}} {(\%)}$}      & 18.24 & 4.37 & 19.22 & 23.61 & 24.18 & 17.92 \\
          & \multicolumn{1}{|c|}{$r_{\text{rel}} (^{\circ})$}     & 2.69  & 1.18 & 3.54  & 4.11  & 2.47  & 2.80 \\
    \hline
    \multirow{2}{*}{ORB-SLAM} & \multicolumn{1}{|c|}{$t_{\text{rel}} {(\%)}$}      & 25.29 & 26.30     & 26.01 & 24.53 & 32.40 & 27.06 \\
          & \multicolumn{1}{|c|}{$r_{\text{rel}} (^{\circ})$}     & 7.37  & 3.10     & 10.62 & 10.83 & 12.13 & 10.24 \\
    \bottomrule
    \end{tabularx}%
    \begin{itemize}
		\scriptsize
		\item $t_{\text{rel}}$: average translational RMSE drift $(\%)$ on length of 100m-800m.
		\item $r_{\text{rel}}$: average rotational RMSE drift ($^{\circ}/100$m) on length of 100m-800m.
	\end{itemize}
  \label{tab:vo_traj}%
\end{table}%

\begin{table}
  \centering
  \caption{
{\color{black}
Comparisons to monocular VIO approaches on KITTI Odometry sequence 10. We present medians, first quartiles, and third quartiles of translational errors in meters. 
}
The results for the benchmark methods are reproduced from \cite{clark2017vinet, shamwell2019unsupervised}. 
We report errors on distances of $100,200,300,400,500$ m from KITTI Odometry sequence 10 to have identical metrics with \cite{clark2017vinet,shamwell2019unsupervised}. Full results for SelfVIO on sequence 10 can be found in Tab. \ref{tab:0910}.}
    \begin{tabular}{lcccccc}
    \toprule
     &  & 100m & 200m    & 300m & 400m & 500m \\
    \hline
    \multirow{3}{*}{SelfVIO} & \multicolumn{1}{|c|}{Med.}     & 1.18  & 2.85  & \textbf{5.11} & \textbf{7.48} & \textbf{8.03} \\
          & \multicolumn{1}{|c|}{1st Quar.}     & 0.82  & 2.03  & 3.09  & 5.31  & 6.59 \\
		  & \multicolumn{1}{|c|}{3rd Quar.}     & 1.77  & 3.89  & 7.15  & 9.26  & 10.29 \\
    \hline
    \multirow{3}{*}{\shortstack[l]{SelfVIO\\(no IMU)}} & \multicolumn{1}{|c|}{Med.}      & 2.25  & 4.3   & 7.29  & 13.11 & 17.29 \\
          & \multicolumn{1}{|c|}{1st Quar.}     & 1.33  & 2.92  & 5.51  & 10.34 & 15.26 \\
		  & \multicolumn{1}{|c|}{3rd Quar.}     & 2.64  & 5.57  & 10.93 & 15.17 & 19.07 \\
    \hline
    \multirow{3}{*}{\shortstack[l]{SelfVIO\\(LSTM)}} & \multicolumn{1}{|c|}{Med.}      & 1.21  & 3.08   & 5.35  & 7.81 & 9.13 \\
          & \multicolumn{1}{|c|}{1st Quar.}     & 0.81  & 2.11  & 3.18  & 5.76 & 6.61 \\
		  & \multicolumn{1}{|c|}{3rd Quar.}     & 1.83  & 4.76  & 8.06 & 9.41 & 10.95 \\
    \hline
    \multirow{3}{*}{VIOLearner} & \multicolumn{1}{|c|}{Med.}      & 1.42  & 3.37  & 5.7   & 8.83  & 10.34 \\
          & \multicolumn{1}{|c|}{1st Quar.}     & 1.01  & 2.27  & 3.24  & 5.99  & 6.67 \\
		  & \multicolumn{1}{|c|}{3rd Quar.}     & 2.01  & 5.71  & 8.31  & 10.86 & 12.92 \\
    \hline
    \multirow{3}{*}{VINET} & \multicolumn{1}{|c|}{Med.}      & \textbf{0} & \textbf{2.5} & 6     & 10.3  & 16.8 \\
          & \multicolumn{1}{|c|}{1st Quar.}     & 0     & 1.01  & 3.26  & 5.43  & 8.6 \\
		  & \multicolumn{1}{|c|}{3rd Quar.}     & 2.18  & 5.43  & 17.9  & 39.6  & 70.1 \\
    \hline
    \multirow{3}{*}{EKF+VISO2} & \multicolumn{1}{|c|}{Med.}      & 2.7   & 11.9  & 26.6  & 40.7  & 57 \\
          & \multicolumn{1}{|c|}{1st Quar.}     & 0.54  & 4.89  & 9.23  & 13    & 19.5 \\
		  & \multicolumn{1}{|c|}{3rd Quar.}     & 9.2   & 32.6  & 58.1  & 83.6  & 98.9 \\
    \bottomrule
    \end{tabular}%
  \label{tab:vio_traj}%
\end{table}%

\begin{table}
  \centering
  \caption{Comparisons to monocular VO and monocular VIO approaches on KITTI test sequences 09 and 10.  $t_{rel} (\%)$ is the average translational error percentage on lengths $100-800$m  and $r_{rel} (^{\circ})$ is the rotational error $(^{\circ}/100m)$ on lengths $100-800$m calculated using the standard KITTI benchmark \cite{geiger2013vision}. 
{\color{black}
No loop closure is performed for ORB-SLAM. We evaluate the monocular versions of the compared methods.
} }
		\label{tab:vio_traj_results}
    \begin{tabularx}{\columnwidth}{lXXX|X}
    \toprule
     &  & Seq.09 & Seq.10    & Mean \\
    \hline
    \multirow{2}{*}{SelfVIO} & \multicolumn{1}{|c|}{$t_{\text{rel}} {(\%)}$}     & \textbf{1.95} & \textbf{1.81} & \textbf{1.88} \\
          & \multicolumn{1}{|c|}{$r_{\text{rel}} (^{\circ})$}     & \textbf{1.15} & \textbf{1.30} & \textbf{1.23} \\
    \hline
    \multirow{2}{*}{SelfVIO (no IMU)} & \multicolumn{1}{|c|}{$t_{\text{rel}} {(\%)}$}      & 2.49  & 2.33 & 2.41  \\
          & \multicolumn{1}{|c|}{$r_{\text{rel}} (^{\circ})$}     & 1.28  & 1.96 & 1.62 \\    
    \hline
    \multirow{2}{*}{SelfVIO (LSTM)} & \multicolumn{1}{|c|}{$t_{\text{rel}} {(\%)}$}      & 2.10  & 2.03 & 2.07  \\
          & \multicolumn{1}{|c|}{$r_{\text{rel}} (^{\circ})$}     & 1.19  & 1.44 & 1.32 \\    
    \hline
    \multirow{2}{*}{VIOLearner} & \multicolumn{1}{|c|}{$t_{\text{rel}} {(\%)}$}      & 2.27  & 2.74 & 2.53  \\
          & \multicolumn{1}{|c|}{$r_{\text{rel}} (^{\circ})$}     & 1.52  & 1.35 & 1.31 \\    
    \hline
    \multirow{2}{*}{SfMLearner} & \multicolumn{1}{|c|}{$t_{\text{rel}} {(\%)}$}      & 21.63 & 20.54 & 21.09\\
          & \multicolumn{1}{|c|}{$r_{\text{rel}} (^{\circ})$}     & 3.57  & 10.93 & 7.25\\
	\hline
    \multirow{2}{*}{Zhan et al.} & \multicolumn{1}{|c|}{$t_{\text{rel}} {(\%)}$}      & 11.92  & 12.62 & 12.27   \\
          & \multicolumn{1}{|c|}{$r_{\text{rel}} (^{\circ})$}     & 3.60  & 3.43 & 3.52 \\
    \hline
    \multirow{2}{*}{ORB-SLAM} & \multicolumn{1}{|c|}{$t_{\text{rel}} {(\%)}$}      & 45.52 & 6.39     & 25.96 \\
          & \multicolumn{1}{|c|}{$r_{\text{rel}} (^{\circ})$}     & 3.10  & 3.20     & 3.15 \\
          \hline
    \multirow{2}{*}{OKVIS} & \multicolumn{1}{|c|}{$t_{\text{rel}} {(\%)}$}      & 9.77 & 17.30     & 13.51 \\
          & \multicolumn{1}{|c|}{$r_{\text{rel}} (^{\circ})$}     & 2.97  & 2.82    & 2.90 \\
          \hline
    \multirow{2}{*}{ROVIO} & \multicolumn{1}{|c|}{$t_{\text{rel}} {(\%)}$}      & 20.18 & 20.04     & 10.11 \\
          & \multicolumn{1}{|c|}{$r_{\text{rel}} (^{\circ})$}     & 2.09  & 2.24    & 2.17 \\
	\midrule
    \multirow{2}{*}{ORB-SLAM$^\dagger$} & \multicolumn{1}{|c|}{$t_{\text{rel}} {(\%)}$}      & 24.41 & 3.16     &  13.79 \\
          & \multicolumn{1}{|c|}{$r_{\text{rel}} (^{\circ})$}     & 2.08  & 2.15    &  2.12\\
          \hline
    \multirow{2}{*}{OKVIS$^\dagger$} & \multicolumn{1}{|c|}{$t_{\text{rel}} {(\%)}$}      & 5.69 & 10.82     &  8.26 \\
          & \multicolumn{1}{|c|}{$r_{\text{rel}} (^{\circ})$}     & 1.89  & 1.80    &  1.85\\
          \hline
    \multirow{2}{*}{ROVIO$^\dagger$} & \multicolumn{1}{|c|}{$t_{\text{rel}} {(\%)}$}      & 12.38 & 10.74    &  11.56 \\
          & \multicolumn{1}{|c|}{$r_{\text{rel}} (^{\circ})$}     & 1.71  & 1.75    &  1.73\\
    \bottomrule
    \end{tabularx}%
    \begin{itemize}
		\scriptsize
		\item $\dagger$: Full resolution input image ($1242\times376$)
		\item $t_{\text{rel}}$: average translational RMSE drift $(\%)$ on length of 100m-800m.
		\item $r_{\text{rel}}$: average rotational RMSE drift ($^{\circ}/100$m) on length of 100m-800m.
	\end{itemize}
  \label{tab:0910}%
\end{table}%

\begin{figure*}
\centering
\includegraphics[width=\textwidth]{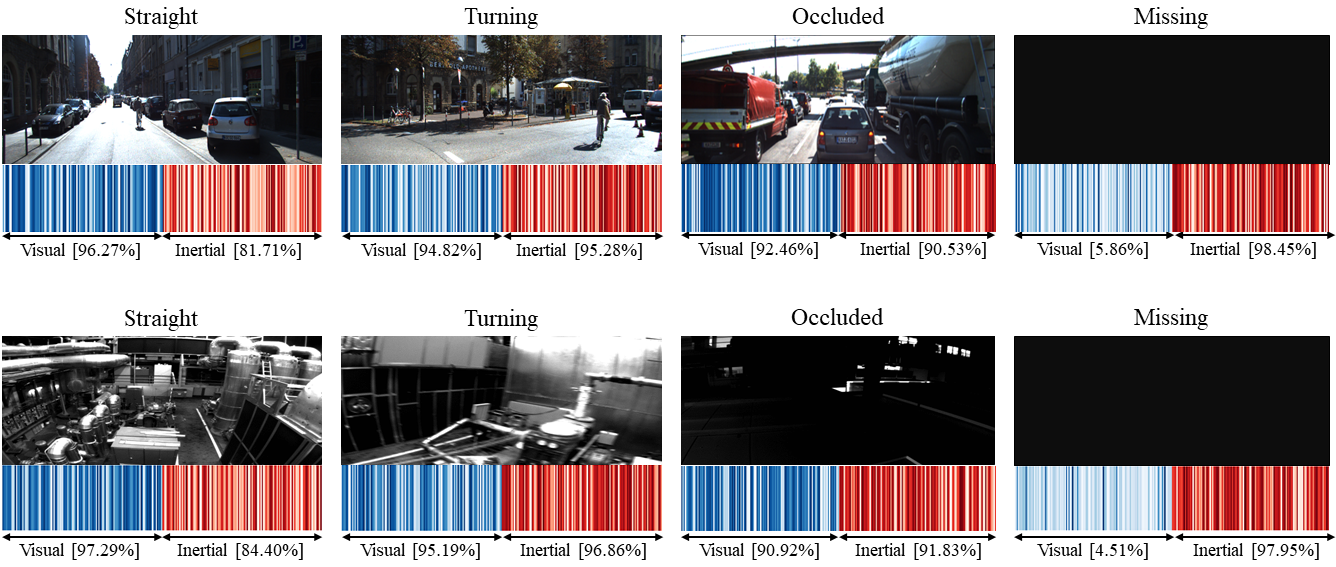}
\caption{
{\color{black}
\textbf{Visualization of the adaptive fusion under different conditions.} The weights and mean percentage activation of visual and inertial features shown at the bottom of each frame reflect the ego-motion dynamics: top: KITTI dataset, bottom: EuRoC dataset. Visual features dominate over inertial features during straight motions, whereas they diminish in importance when faced with a lack of salient visual signals. In the cases of turning and occlusion, the importance of inertial features increases to compensate for the lost visual features due to the reduced overlap between the consecutive frames.
}}
\label{fig:attention-vis}
\end{figure*}

\subsection{Network Training}
We implement the architecture with the publicly available Tensorflow framework \cite{abadi2016tensorflow}. Batch normalization is employed for all of the layers except for the output layers.
Three consecutive images are stacked together to form the input batch, where the central frame is the target view for the depth
estimation.
We augment the data with random scaling, cropping and horizontal flips.
SelfVIO is trained for $100,000$ iterations using a batch size
of $16$. During the network training, we calculate error on the
validation set at intervals of $1,000$ iterations. 
We use the ADAM~\cite{kingma2014adam}
solver with $momentum1=0.9$, $momentum2=0.99$, $gamma=0.5$,
learning rate=$2e-4$, and an exponential learning rate policy. The
network is trained using single-point precision (FP32) on a
desktop computer with a 3.00 GHz Intel i7-6950X processor and
NVIDIA Titan V GPUs. 
{\color{black}
The proposed model runs at $81$ ms per frame on a Titan V GPU, taking $33$ ms for depth generation, $27$ ms for visual odometry, and $21$ ms for IMU processing and sensor fusion.
}

\subsection{Evaluation}
\label{subsec:evaluation}
We compare our approach to a collection of recent VO, VIO, and
VSLAM approaches described earlier in Section \ref{sec:rel_work}:
\begin{itemize}
\item Learning-based methods:
	\begin{itemize}
	\item SFMLearner \cite{zhou2017unsupervised}
	\item Mahjourian et al. \cite{mahjourian2018unsupervised} (results reproduced from \cite{mahjourian2018unsupervised})
	\item Zhan et al. \cite{zhan2018unsupervised} (results reproduced from \cite{zhan2018unsupervised})
	\item VINet \cite{clark2017vinet}
	\item UnDeepVO \cite{li2018undeepvo}
	\item Geonet \cite{yin2018geonet}
	\item DF-Net \cite{zou2018df}
	\item Competitive Collaboration (CC) \cite{ranjan2019competitive}
	\item VIOLearner-RGB \cite{shamwell2019unsupervised}
	\end{itemize}
\item Traditional methods:
	\begin{itemize}
	\item OKVIS \cite{leutenegger2015keyframe}
	\item ROVIO \cite{bloesch2017iterated}
	\item VISO2 (results reproduced from \cite{shamwell2019unsupervised})
	\item ORB-SLAM (results reproduced from \cite{shamwell2019unsupervised})
	\item SVO+MSF \cite{faessler2016autonomous, forster2016svo}
	\item VINS-Mono \cite{qin2018vins}
	\item EKF+VISO2 (results reproduced from \cite{clark2017vinet})
	\item MSCKF \cite{mourikis2007multi}	
	\end{itemize}
\end{itemize}

We include monocular versions of competing algorithms to have a common setup with our method.
SFMLearner, Mahjourian et al., Zhan
et al., and VINet optimize over multiple consecutive monocular images or stereo image pairs; and OKVIS and ORB-SLAM perform bundle adjustment.
Similarly, we include the RGB version of VIOLearner for all the comparisons, which uses RGB image input and the monocular depth generation sub-network from SFMLearner \cite{shamwell2019unsupervised} rather than RGB-depth data. 
We perform 6-DOF least-squares Umeyama alignment \cite{umeyama1991least} for
trajectory alignment on monocular approaches as they lack scale information.
For SFMLearner, we follow \cite{zhou2017unsupervised} to estimate the scale from the ground
truth for each estimate. 
We evaluate the compared methods at images scaled down to size $256\times 832$ to match the image resolution used by SelfVIO.

We train separate networks for KITTI
and EuRoC datasets for benchmarking and the Cityscapes dataset \cite{cordts2016cityscapes} for evaluating the cross-dataset generalization ability of the model. SelfVIO implicitly learns to estimate camera-IMU extrinsics and IMU instrinsics directly from raw data, enabling
SelfVIO to translate from one
dataset (with a given camera-IMU configuration) to another (with
a different camera-IMU configuration). 

\begin{table}
\centering
\caption{Absolute Trajectory Error (ATE) in meters on KITTI odometry dataset. We also report the results of the other methods for comparison that are taken from \cite{zhou2017unsupervised, yin2018geonet}. Our method outperforms all of the compared methods. {\color{black}
No loop closure is performed for ORB-SLAM. 
}}
\begin{tabularx}{\columnwidth}{X|X|X}
\toprule
Method & Seq.09 & Seq.10 \\
\hline
ORB-SLAM~(full)& $0.014\pm 0.008$ & $0.012\pm 0.011$\\
ORB-SLAM~(short) & $0.064\pm 0.141$ & $0.064\pm 0.130$\\
Zhou\cite{zhou2017unsupervised} & $0.021\pm 0.017$ & $0.020\pm 0.015$\\
SfM-Learner \cite{zhou2017unsupervised} & $0.016\pm 0.009$ & $0.013\pm 0.009$\\
GeoNet~\cite{yin2018geonet} &0.012 $\pm$ 0.007 & 0.012 $\pm$ 0.009\\
CC~\cite{ranjan2019competitive} & 0.012 $\pm$ 0.007 & 0.012 $\pm$ 0.008  \\
GANVO~\cite{almalioglu2019ganvo} &0.009 $\pm$ 0.005 & 0.010 $\pm$ 0.013\\
VIOLearner \cite{shamwell2019unsupervised} & 0.012 & 0.012 \\
SelfVIO (ours) &\bf{ 0.008 $\pm$ 0.006 }& \bf{0.009 $\pm$ 0.008}\\
\bottomrule
\end{tabularx}
\label{tab:pose_snippet}
\end{table}

\subsubsection{Ablation Studies}
We perform two ablation studies on our proposed network and call these SelfVIO (no IMU) and SelfVIO (LSTM).
\paragraph{Visual Vs. Visual-Inertial} 
We disable the inertial odometry module and omit IMU data; instead, we use a vision-only odometry to
estimate the initial warp.
This version of the network is referred to as SelfVIO (no IMU)
and results are only included to provide additional perspective on
the vision-only performance of our architecture (and specifically
the adversarial training) compared to other vision-only approaches.

\paragraph{CNN vs RNN} 
Additionally, we perform ablation studies where we replace the convolutional network described in Sec. \ref{sec:inertial_odometry} with a recurrent neural network, specifically a bidirectional LSTM to process IMU input at the cost of an increase in the number of parameters and, hence, more computational power.
This version of the network is referred to as SelfVIO (LSTM).

\subsubsection{Spatial Misalignments}
We test the robustness of our method against camera-sensor miscalibration. We introduce calibration errors by adding a rotation of a chosen magnitude and random angle $\Delta R_{s} \sim \mbox{vMF}(\cdot|\mu,\kappa)$ to the camera-IMU rotation matrices $R_{s}$, where $\mbox{vMF}(\cdot|\mu,\kappa)$ is the von Mises-Fisher distribution \cite{wood1994simulation}, $\mu$ is the directional mean and $\kappa$ is the concentration parameter of the distribution. 
{\color{black}
We apply the calibration offsets during testing. Note that these are never used during training.
}

\begin{figure}
	\begin{subfigure}{\columnwidth} 
	    \includegraphics[width=1.0\columnwidth]{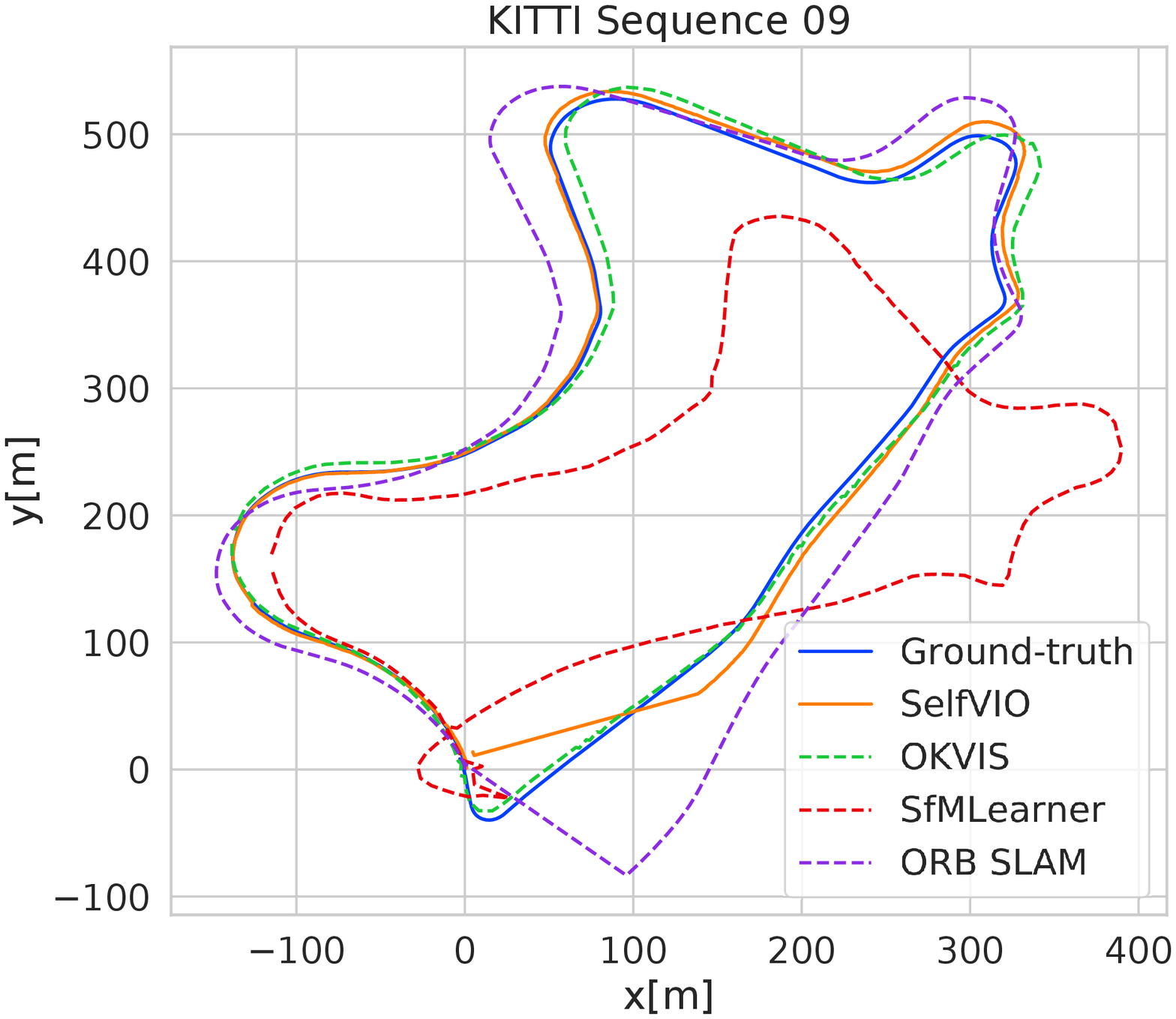}
        \label{fig:seq09}
	\end{subfigure}
	\begin{subfigure}{\columnwidth} 
	    \includegraphics[width=1.0\columnwidth]{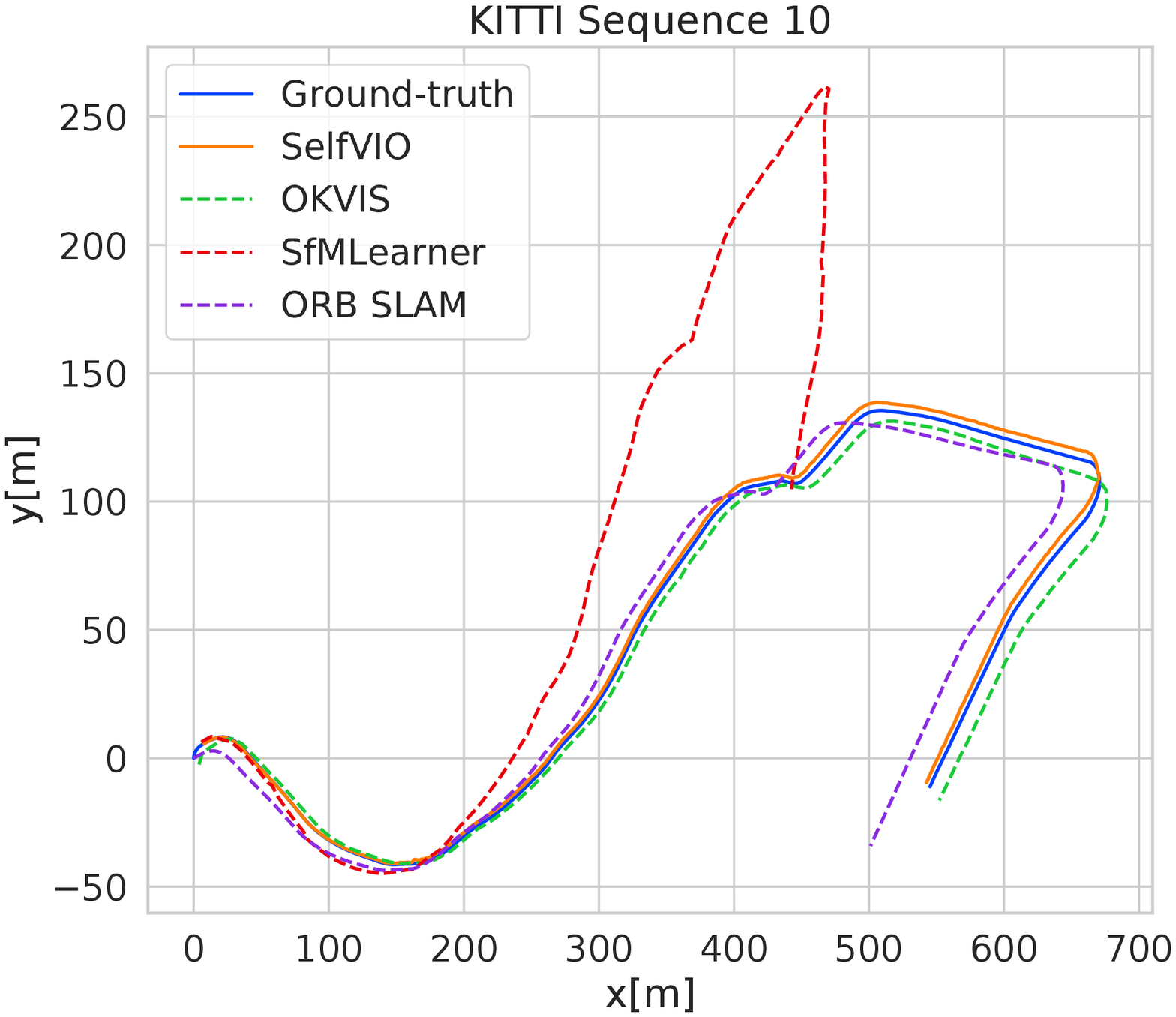}
        \label{fig:seq10}
	\end{subfigure}	
	\caption{
{\color{black}
Sample trajectories comparing the proposed unsupervised learning method SelfVIO with monocular versions of ORB SLAM, OKVIS, SfMLearner, and the ground truth in meter scale on KITTI sequences 09 and 10.
} SelfVIO shows a better odometry estimation in terms of both rotational and translational motions. }
    \label{fig:traj_kitti_results}
\end{figure}

\subsubsection{Evaluation Metrics}
We evaluate our trajectories primarily using the standard KITTI
relative error metric (reproduced below from \cite{geiger2013vision}):
\begin{equation}
E_{rot}(\mathcal{F}) = \frac{1}{\vert \mathcal{F} \vert} \sum_{(i,j)\in \mathcal{F}} \Vert (\hat{\mathbf{q}}_i \ominus \hat{\mathbf{q}}_j )\ominus( \mathbf{q}_i \ominus \mathbf{q}_j ) \Vert_2 ,	
\end{equation}
\begin{equation}
E_{trans}(\mathcal{F}) = \frac{1}{\vert \mathcal{F} \vert} \sum_{(i,j)\in \mathcal{F}} \Vert (\hat{\mathbf{p}}_i \ominus \hat{\mathbf{p}}_j )\ominus( \mathbf{p}_i \ominus \mathbf{p}_j ) \Vert_2 ,	
\end{equation}
where $\mathcal{F}$ is a set of frames, $\ominus$ is the inverse compositional operator, $\mathbf{x}=[\mathbf{p}, \mathbf{q}] \in SE(3)$ and $\hat{\mathbf{x}}=[\hat{\mathbf{p}}, \hat{\mathbf{q}}] \in SE(3)$ are estimated and true pose values as elements of Lie group $SE(3)$, respectively.

For KITTI dataset, we also evaluate the errors at lengths of $100, 200, 300, 400,$ and $500$ m. 
Additionally, we compute the root mean squared error (RMSE)
for trajectory estimates on five frame snippets as has been
done recently in \cite{zhou2017unsupervised, mahjourian2018unsupervised}.

We evaluate depth estimation performance of each method using several error and accuracy metrics from prior works \cite{eigen2014depth}:
{
\newline
\begin{tabular}{l}

Threshold:  \% of $y_{i}$ s.t. $\max(\frac{y_i}{y_i^*},\frac{y_i^*}{y_i}) = \delta < thr$
\\ 
RMSE (linear):  $\sqrt{\frac1{|T|}\sum_{y\in T}||y_i - y_i^*||^2}$
\\
Abs relative difference:  $\frac1{|T|}\sum_{y\in T}|y - y^*| / y^*$
\\ 
RMSE (log):  $\sqrt{\frac1{|T|}\sum_{y\in T}||\log y_i - \log y_i^*||^2}$ 
\\
Squared relative difference:  $\frac1{|T|}\sum_{y\in T}||y - y^*||^2 / y^*$
\\ 

\end{tabular}
}

\begin{table*}  
  \caption{Absolute translation errors (RMSE) in meters for all trials in the EuRoC MAV dataset, 
  {\color{black} 
  using monocular versions of all the compared methods.
  } Errors have been computed after the estimated trajectories were aligned with the
ground-truth trajectory using the method in \cite{umeyama1991least}. The top performing algorithm on each platform and dataset is highlighted in bold.}
    \begin{tabularx}{\textwidth}{l|X|X|X|X|X|X|X|X|X|X|X}
	\toprule
          & MH01 (E) & MH02 (E)$\ddagger$ & MH03 (M) & MH04 (D)$\ddagger$ & MH05 (D) &V101 (E) & V102 (M) & V103 (D)$\ddagger$ & V201 (E) & V202 (M)$\ddagger$ & V203 (D) \\
	\hline
    OKVIS & 0.23  & 0.30  & 0.33  & 0.44  & 0.59  & 0.12  & 0.26  & 0.33  & 0.17  & 0.21  & 0.37 \\
    ROVIO & 0.29  & 0.33  & 0.35  & 0.63  & 0.69  & 0.13  & 0.13  & 0.18  & 0.17  & 0.19  & 0.19 \\
    VINSMONO & 0.35  & 0.21  & 0.24  & 0.29  & 0.46  & 0.10  & 0.14  & 0.17  & 0.11  & 0.12  & 0.26 \\
    SelfVIO & 0.19 & 0.15 & 0.21  & \textbf{0.16} & \textbf{0.29}  & 0.08 & \textbf{0.09} & \textbf{0.10} & 0.11 & \textbf{0.08} & \textbf{0.11} \\
	\midrule
	SVOMSF$^\dagger$ & 0.14  & 0.20  & 0.48  & 1.38  & 0.51  & 0.40  & 0.63  & x & 0.20  & 0.37  & x \\
    MSCKF$^\dagger$ & 0.42  & 0.45  & 0.23  & 0.37  & 0.48  & 0.34  & 0.20  & 0.67  & 0.10  & 0.16  & 1.13 \\
    OKVIS$^\dagger$ & \textbf{0.16}  & 0.22  & 0.24  & 0.34  & 0.47  & 0.09  & 0.20  & 0.24  & 0.13  & 0.16  & 0.29 \\
    ROVIO$^\dagger$ & 0.21  & 0.25  & 0.25  & 0.49  & 0.52  & 0.10  & 0.10  & 0.14  & 0.12  & 0.14  & 0.14 \\
    VINSMONO$^\dagger$ & 0.27  & \textbf{0.12}  & \textbf{0.13}  & 0.23  & 0.35  & \textbf{0.07}  & 0.10  & 0.13  & \textbf{0.08}  & \textbf{0.08}  & 0.21 \\
	\bottomrule
    \end{tabularx}%
    {\color{black}
    \begin{itemize}
		\scriptsize
		\item $\dagger$: Full resolution input image ($752x480$)
		\item $\ddagger$: EuRoC test sequences
	\end{itemize}
	}
  \label{tab:euroc_vio}%
\end{table*}%

\begin{table}
  \centering
  \caption{
{\color{black}
Effectiveness of the compared methods in the presence of miscalibrated input. We report the relative rotational and translational errors for (a) temporal and (b) translational offsets spanning a range of several orders of magnitude on Euroc dataset, using monocular versions of the compared methods. 
}}
	\begin{subtable}{\columnwidth}
		\begin{tabular}{lcccc}
		\toprule
		 &  & \multicolumn{3}{c}{Translational offset (m)} \\
		 &  & 0.05 & 0.15 & 0.30 \\
		\hline
		\multirow{2}{*}{OKVIS} & \multicolumn{1}{|c|}{$t_{\text{rel}} {(\%)}$}      & $5.21\pm1.95$ & $20.39\pm5.06$     & $71.53\pm15.92$ \\
			  & \multicolumn{1}{|c|}{$r_{\text{rel}} (^{\circ})$}     & $5.16\pm1.15$  & $14.54\pm2.41$    & $65.48\pm5.62$  \\
			  \hline
		\multirow{2}{1cm}{VINS-Mono} & \multicolumn{1}{|c|}{$t_{\text{rel}} {(\%)}$}      & $2.63\pm1.07$ & $15.28\pm4.13$     & $32.47\pm8.86$ \\
			  & \multicolumn{1}{|c|}{$r_{\text{rel}} (^{\circ})$}     & $8.45\pm1.57$  & $18.14\pm2.71$    & $71.31\pm7.49$  \\
		\hline
		\multirow{2}{*}{SelfVIO} & \multicolumn{1}{|c|}{$t_{\text{rel}} {(\%)}$}      & $1.68\pm1.14$  & $10.72\pm3.93$ & $25.18\pm4.35$  \\
			  & \multicolumn{1}{|c|}{$r_{\text{rel}} (^{\circ})$}     & $2.53\pm1.05$  & $9.21\pm2.13$    & $51.37\pm3.71$  \\    
		\bottomrule
		\end{tabular}%
		\caption{}
	\end{subtable}
	\begin{subtable}{\columnwidth}
		\begin{tabular}{lcccc}
			\toprule
			 &  & \multicolumn{3}{c}{Temporal offset (ms)}  \\
			 &  & 15 & 30 & 60 \\
			\hline
			\multirow{2}{*}{OKVIS} & \multicolumn{1}{|c|}{$t_{\text{rel}} {(\%)}$}      & $18.68\pm2.78$ & $24.03\pm4.89$     & $59.73\pm10.87$ \\
				  & \multicolumn{1}{|c|}{$r_{\text{rel}} (^{\circ})$}     & $7.72\pm1.46$  & $18.67\pm3.79$    & $84.63\pm6.59$  \\
				  \hline
			\multirow{2}{1cm}{VINS-Mono} & \multicolumn{1}{|c|}{$t_{\text{rel}} {(\%)}$}      & $10.42\pm1.63$ & $18.81\pm3.19$     & $24.51\pm6.51$ \\
				  & \multicolumn{1}{|c|}{$r_{\text{rel}} (^{\circ})$}     & $9.23\pm1.87$  & $22.19\pm3.27$    & $90.27\pm8.61$  \\
			\hline
			\multirow{2}{*}{SelfVIO} & \multicolumn{1}{|c|}{$t_{\text{rel}} {(\%)}$}      & $5.43\pm1.08$  & $14.31\pm3.57$ & $19.37\pm5.16$ \\
				  & \multicolumn{1}{|c|}{$r_{\text{rel}} (^{\circ})$}     & $3.63\pm1.10$  & $13.29\pm2.85$    & $56.26\pm4.71$  \\    
			\bottomrule
		\end{tabular}%
		\caption{}
	\end{subtable}
    \begin{itemize}
		\scriptsize
		\item $t_{\text{rel}}$: average translational RMSE drift $(\%)$ on length of 100m-800m.
		\item $r_{\text{rel}}$: average rotational RMSE drift ($^{\circ}/100$m) on length of 100m-800m.
	\end{itemize}
  \label{tab:euroc_calib}%
\end{table}%

\section{Results and Discussion}
\label{sec:res_discussion}

In this section, we critically analyse and comparatively discuss our qualitative and quantitative results for depth and motion estimation.

\begin{figure}
\includegraphics[width=\columnwidth]{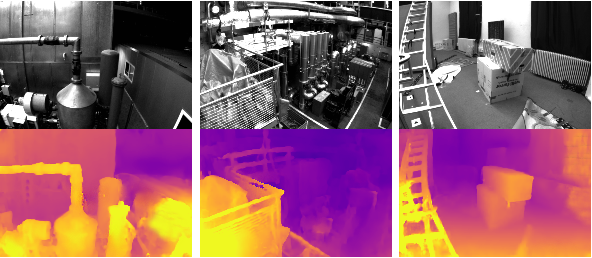}
\caption{
{\color{black}
\textbf{Qualitative results for monocular depth map estimation on the EuRoC dataset.} MAV frames and the corresponding depth maps reconstructed by SelfVIO.
}}
\label{fig:res_depth_euroc}
\end{figure}

\subsection{Monocular Depth Estimation}
We obtain state-of-the-art results on single-view depth prediction as quantitatively shown in Table \ref{tab:depth}. The depth reconstruction performance is evaluated on the Eigen et al.~\cite{eigen2014depth} split of the raw KITTI dataset \cite{geiger2012we}, which is consistent with previous work \cite{eigen2014depth, liu2016learning, mahjourian2018unsupervised, yin2018geonet}. 
All depth maps are capped at 80 meters.
The predicted depth map, $D_{p}$, is multiplied by a scaling factor, $\hat{s}$, that matches the median with the ground truth  depth map, $D_{g}$, to solve the scale ambiguity issue, i.e. $\hat{s} = median(D_g) / median(D_p)$.

Figure \ref{fig:res_depth} shows examples of reconstructed depth maps by the proposed method, GeoNet \cite{yin2018geonet} and the Competitive Collaboration (CC) \cite{ranjan2019competitive}. It is clearly seen that SelfVIO outputs sharper and more accurate depth maps compared to the other methods that fundamentally use an encoder-decoder network with various implementations. An explanation for this result is that adversarial training using the convolutional domain-related feature set of the discriminator distinguishes reconstructed images from the real images, leading to less blurry results \cite{dosovitskiy2016generating}.
{\color{black}
Moreover, although GeoNet \cite{yin2018geonet} and CC \cite{ranjan2019competitive} benchmark methods train additional networks to segment and mask inconsistent regions in the reconstructed frame caused by moving objects, occlusions and re-projection errors, SelfVIO implicitly accounts for these inconsistencies without any need for an additional network.
}
Furthermore, Fig. \ref{fig:res_depth} further implies that the depth reconstruction module proposed by SelfVIO is capable of capturing small objects  in the scene whereas the other methods tend to ignore them. 
A loss function in the image space leads to smoothing out all likely detail locations, whereas an adversarial loss function in feature space with a natural image prior makes the proposed SelfVIO more sensitive to details in the scene \cite{dosovitskiy2016generating}. 
The proposed SelfVIO also performs better in low-textured areas caused by the shading inconsistencies in a scene and predicts the depth values of the corresponding objects much better in such cases.
In Fig. \ref{fig:res_depth_degredation}, we demonstrate typical performance degradation of the compared unsupervised methods that is caused by challenges such as poor road signs in rural areas and huge objects occluding the most of the visual input. Even in these cases, SelfVIO performs slightly better than the existing methods. 

{\color{black}
Moreover, we select a challenging evaluation protocol to test the adaptability of the proposed approach by training on the Cityscapes dataset and fine-tuning on the KITTI dataset (cs+k in Table \ref{tab:depth}).
Although SelfVIO learns the inter-sensor calibration parameters as part of the training process, it can adapt to the new test environment with fine-tuning. 
As the Cityscapes dataset is an RGB-depth dataset, we remove the inertial odometry part and perform an ablation study (SelfVIO (no IMU)) on depth estimation.
While all the learning-based methods in comparison exhibit performance drop in the fine-tuning setting, the results shown in Table \ref{tab:depth} show a clear
advantage of fine-tuning on data that is related to the test set.
}
In this mode (SelfVIO (no IMU)), our network architecture for depth estimation is most similar to GANVO \cite{almalioglu2019ganvo}. 
However, the shared features among the encoder and generator networks enable the network to also have access to low-level information. In addition, the PatchGAN structure in SelfVIO restricts the discriminator from capturing high-frequency structure in depth map estimation.
We observe that using the SelfVIO framework with inertial odometry results in larger performance gains even when it is trained on the KITTI dataset only. 

{\color{black}
Figure \ref{fig:res_depth_euroc} visualizes sample depth maps reconstructed from MAV frames in the EuRoC dataset. Although there is no ground-truth depth map of the frames available in the EuRoC dataset for quantitative analysis, qualitative results in  Fig. \ref{fig:res_depth_euroc} indicates the effectiveness of the depth map reconstruction as well as the efficacy of the proposed approach in datasets containing diverse 6-DoF motions.
}

\begin{figure}
	\begin{subfigure}{\columnwidth} 
	    \includegraphics[width=1.0\columnwidth]{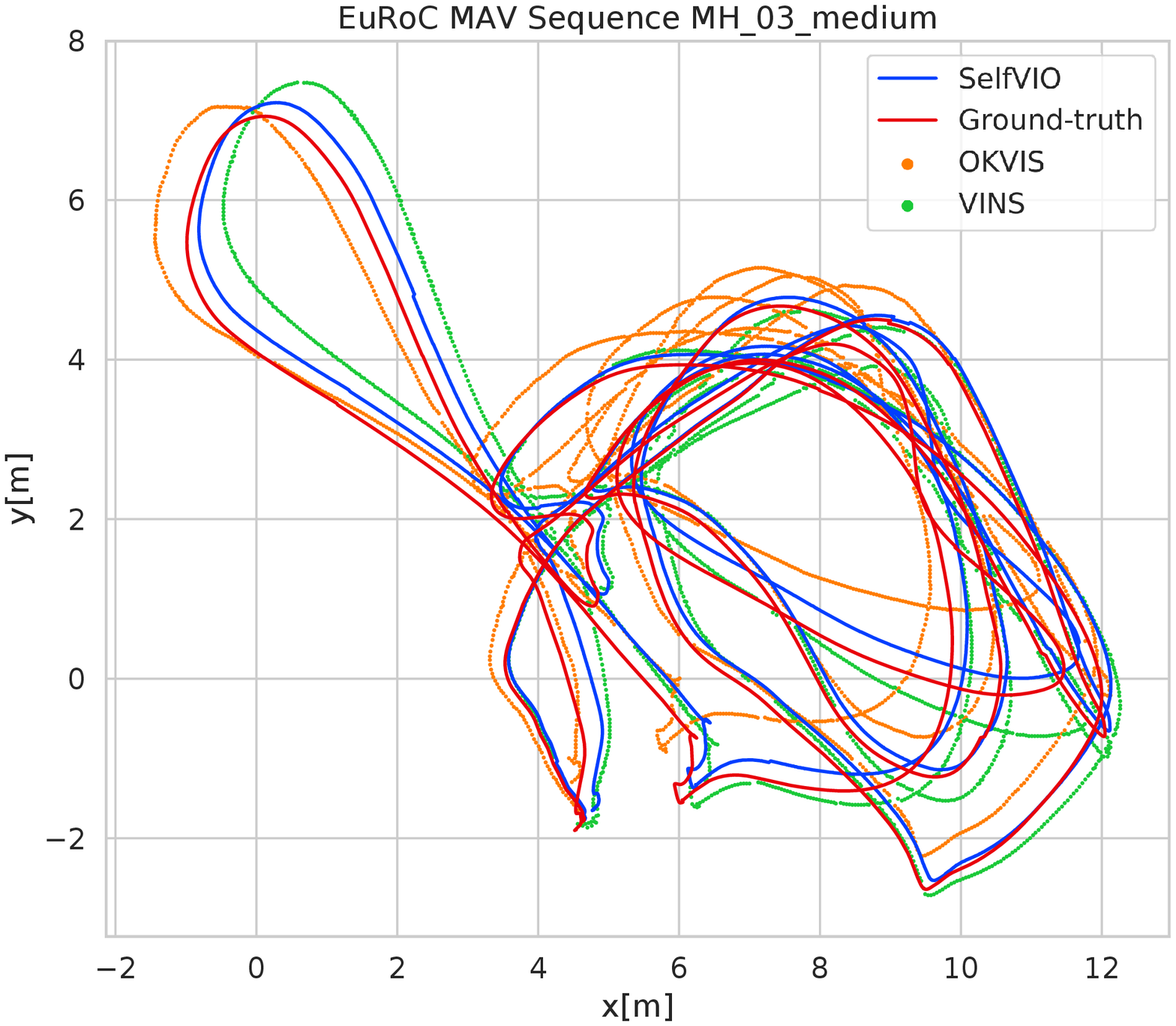}
        \label{fig:seq_mh_03}
	\end{subfigure}
	\begin{subfigure}{\columnwidth} 
	    \includegraphics[width=1.0\columnwidth]{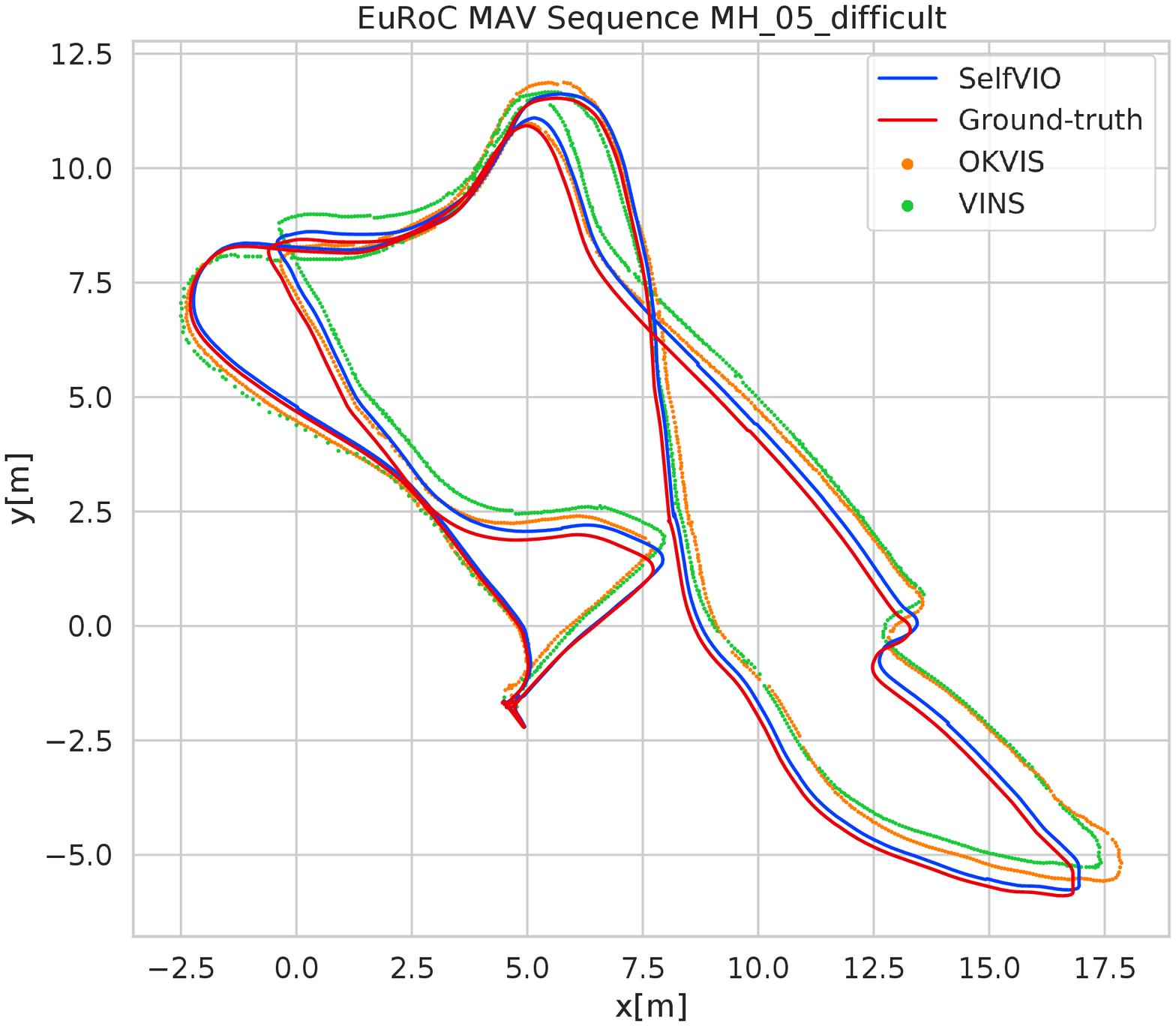}
        \label{fig:seq_mh_05}
	\end{subfigure}	
	\caption{
{\color{black}
Sample trajectories comparing the proposed unsupervised learning method SelfVIO with monocular OKVIS and VINS
}, and the ground truth in meter scale on MH\_03 and MH\_05 sequences of EuRoC dataset. SelfVIO shows a better odometry estimation in terms of both rotational and translational motions. }
    \label{fig:traj_euroc_results}
\end{figure}

\begin{figure*}
	\begin{subfigure}{0.5\textwidth} 
	    \includegraphics[width=1.0\columnwidth]{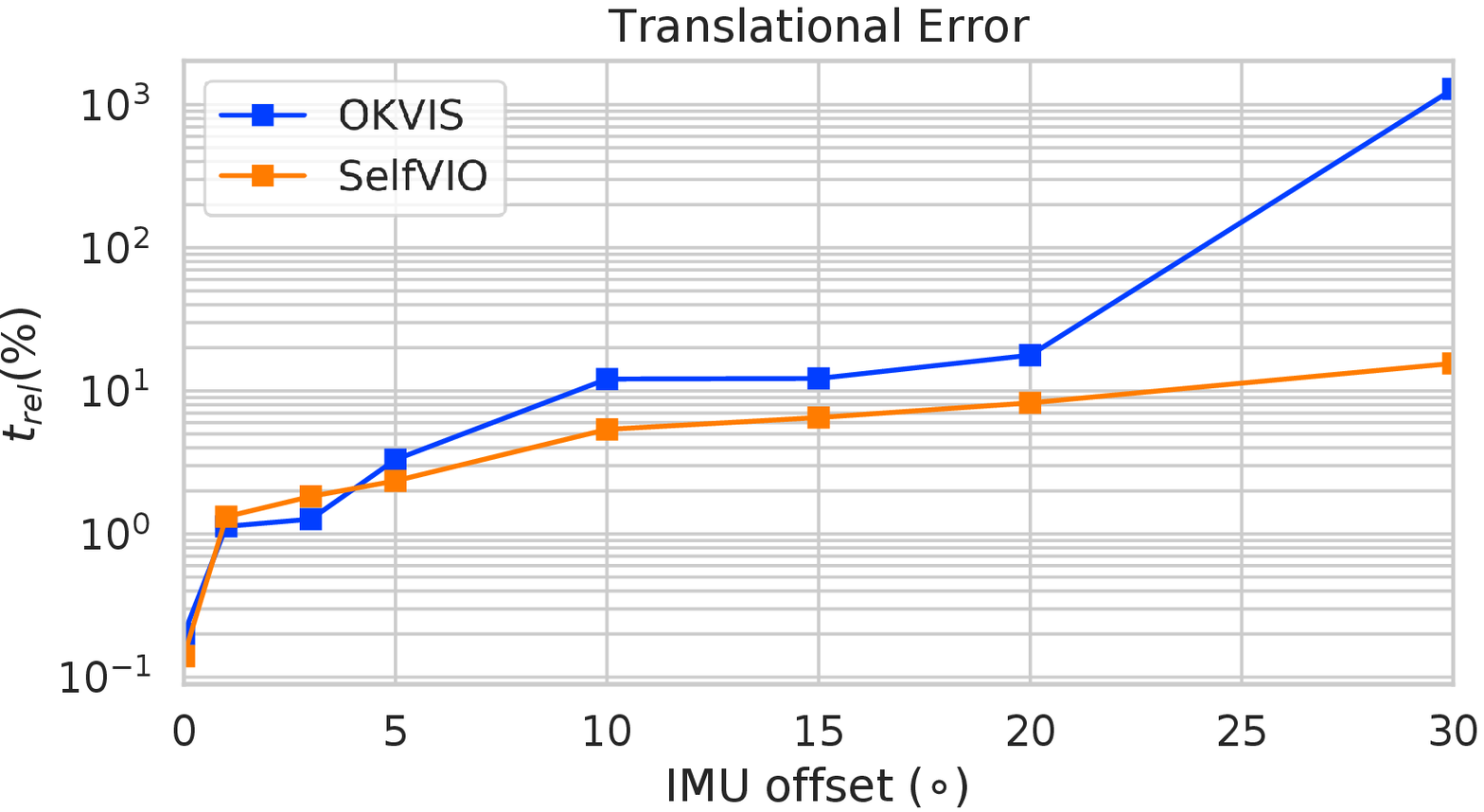}
        \caption{Seq. MH\_03 - Medium (EuRoC)} 
        \label{fig:offset_mh_03_trans}
	\end{subfigure}
	\begin{subfigure}{0.5\textwidth} 
	    \includegraphics[width=1.0\columnwidth]{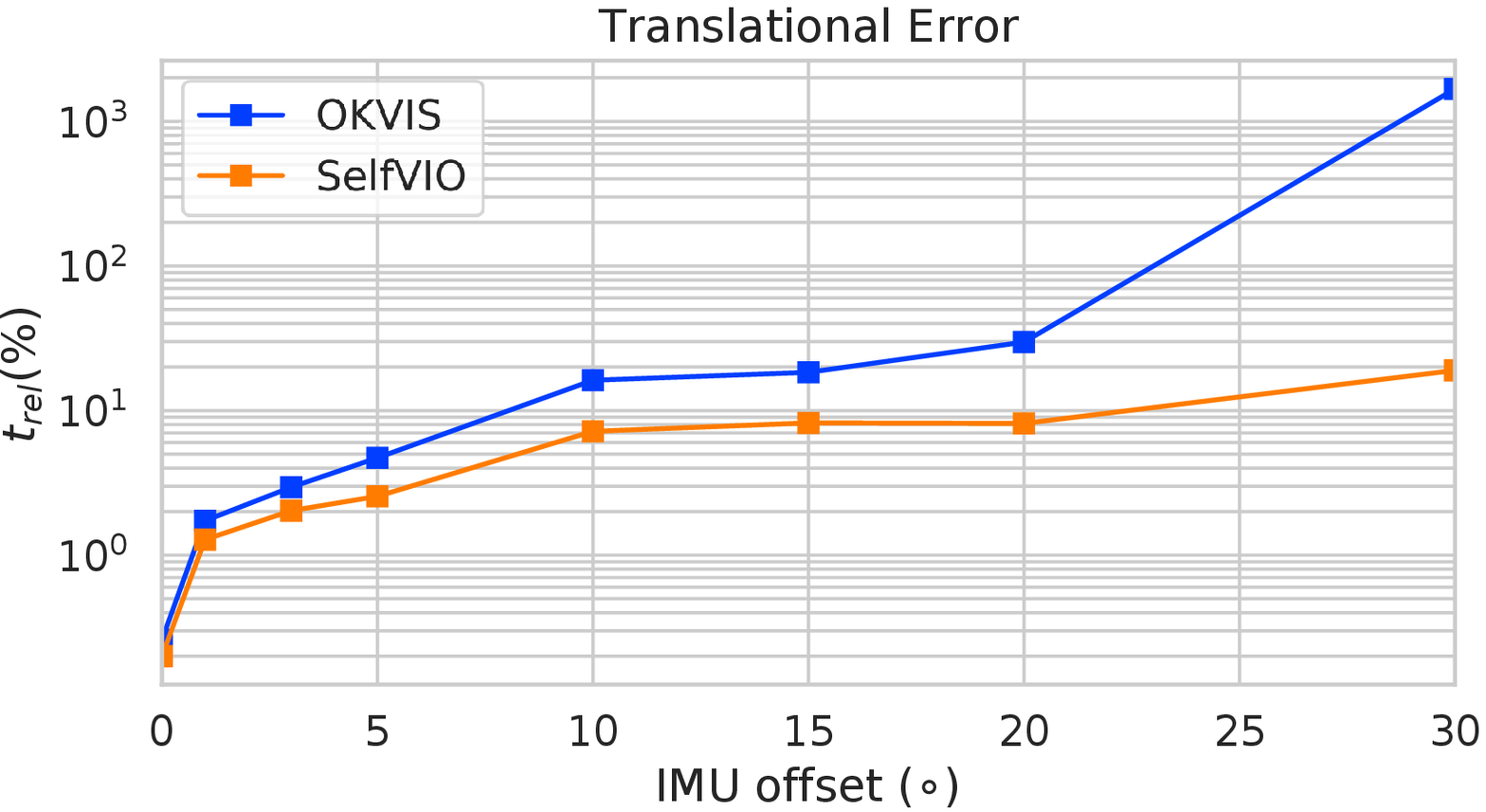}
        \caption{Seq. MH\_05 - Difficult (EuRoC)} 
        \label{fig:offset_mh_05_trans}
	\end{subfigure}
	\par\bigskip 
	\begin{subfigure}{0.5\textwidth} 
	    \includegraphics[width=1.0\columnwidth]{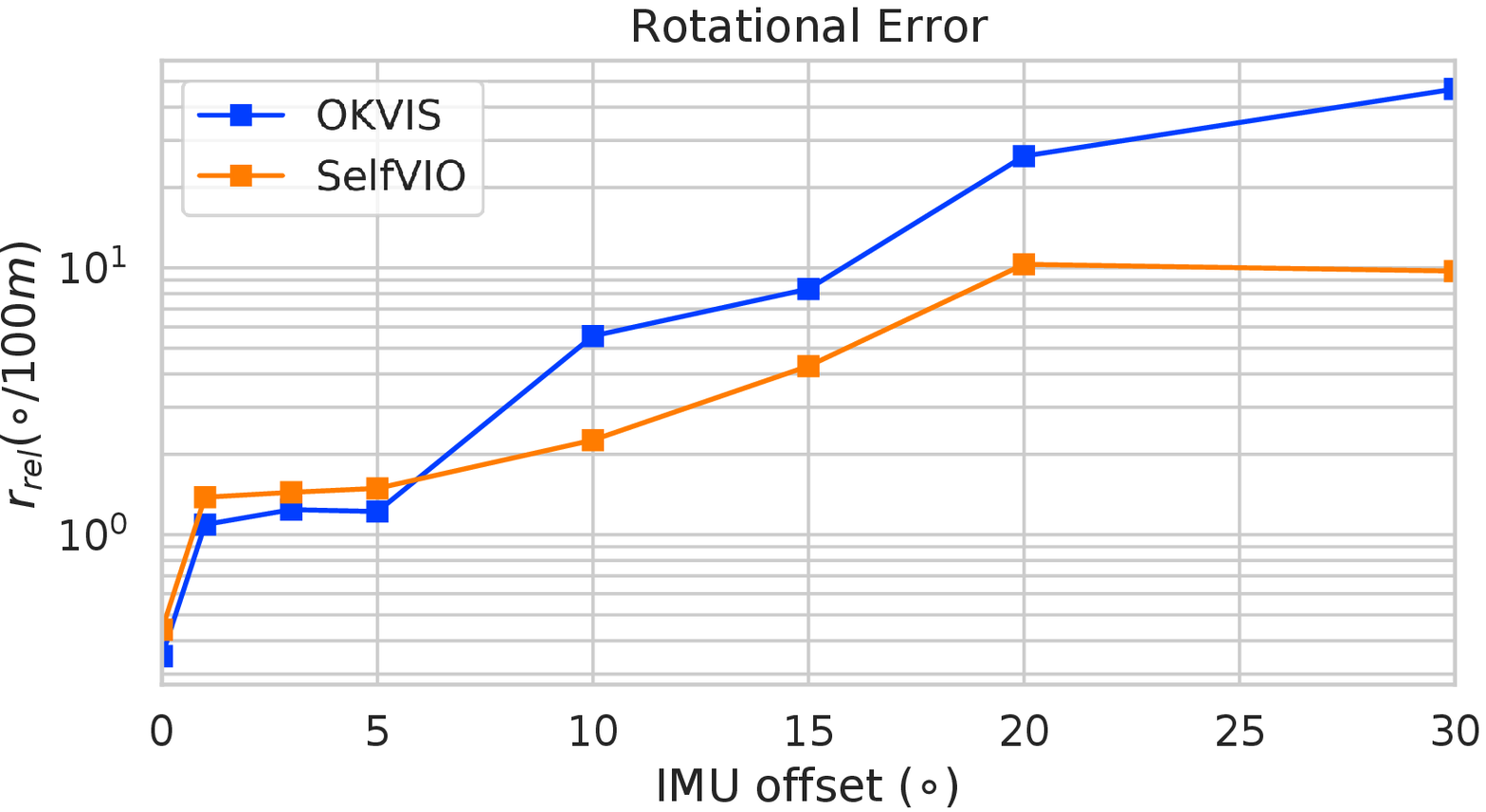}
        \caption{Seq. MH\_03 - Medium (EuRoC)} 
        \label{fig:offset_mh_03_rot}
	\end{subfigure}
	\begin{subfigure}{0.5\textwidth} 
	    \includegraphics[width=1.0\columnwidth]{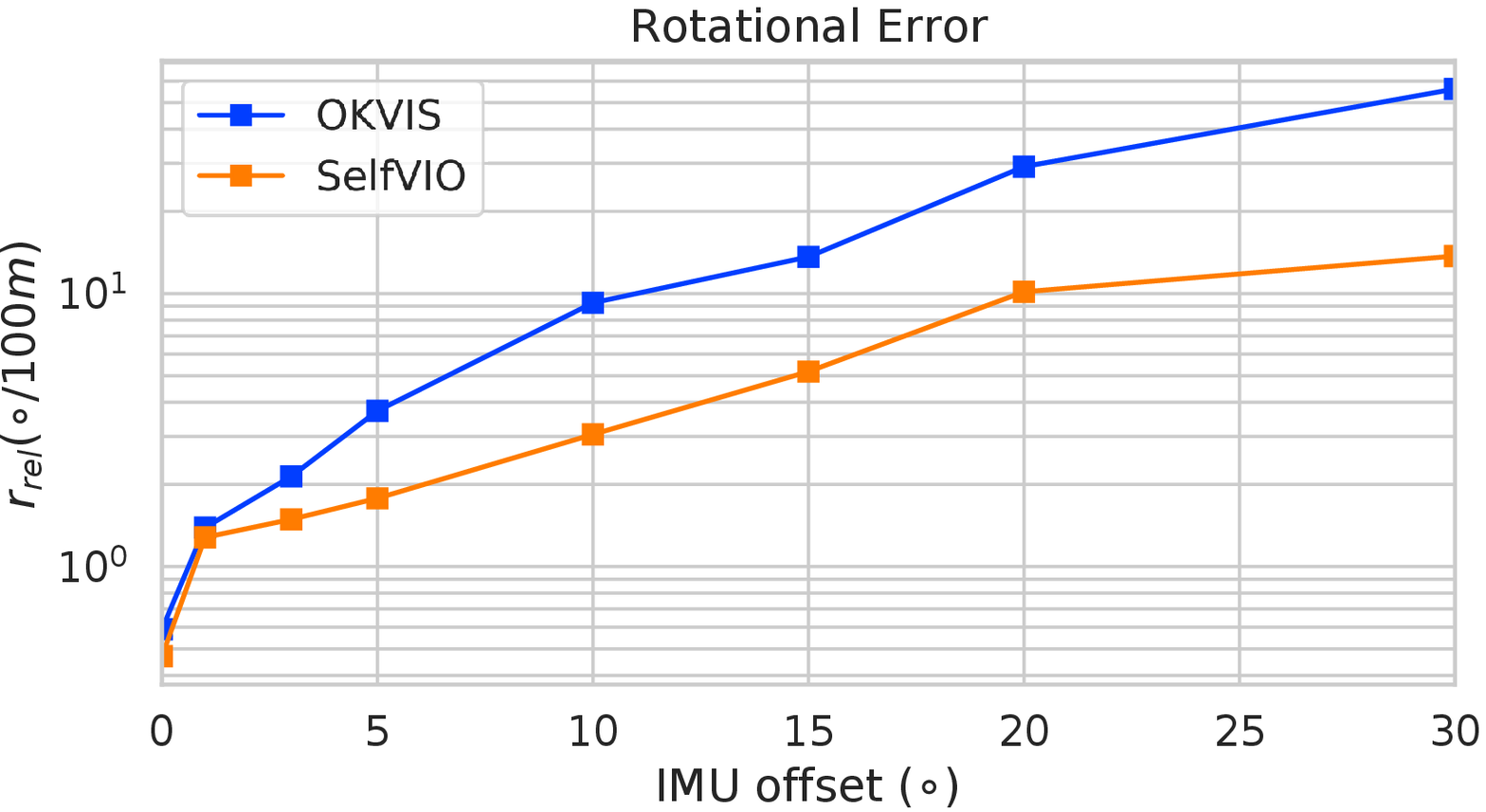}
        \caption{Seq. MH\_05 - Difficult (EuRoC)} 
        \label{fig:offset_mh_05_rot}
	\end{subfigure}		
	\caption{
{\color{black}
Results on SelfVIO and monocular OKVIS trajectory estimation
} on the EuRoC sequences MH\_03 (left column) and MH\_05 (right column) given the induced IMU orientation offset. Measurement errors are shown for each sequence with translational error percentage (top row) 
and rotational error in degrees per $100$m (bottom row) on lengths $25-100$m. In contrast to SelfVIO, after $20-30\deg$, OKVIS exhibits catastrophic failure in translation and orientation estimation. }
    \label{fig:offset_euroc_results}
\end{figure*}

\begin{figure}
	\begin{subfigure}{\columnwidth} 
	    \includegraphics[width=\linewidth]{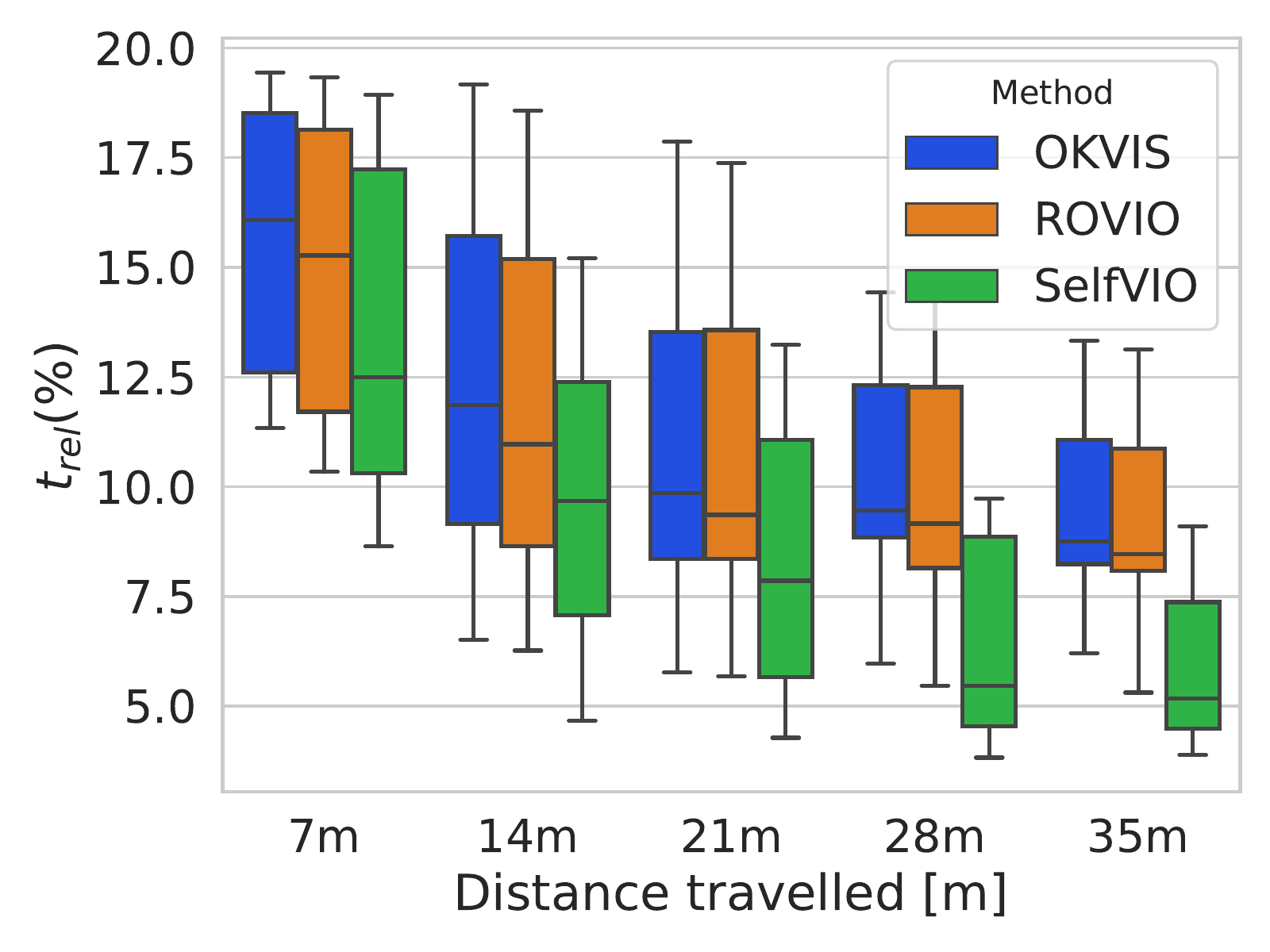}
        \caption{Translational relative pose error} 
        \label{fig:euroc_rpe_trans}
	\end{subfigure}
	\begin{subfigure}{\columnwidth} 
	    \includegraphics[width=\linewidth]{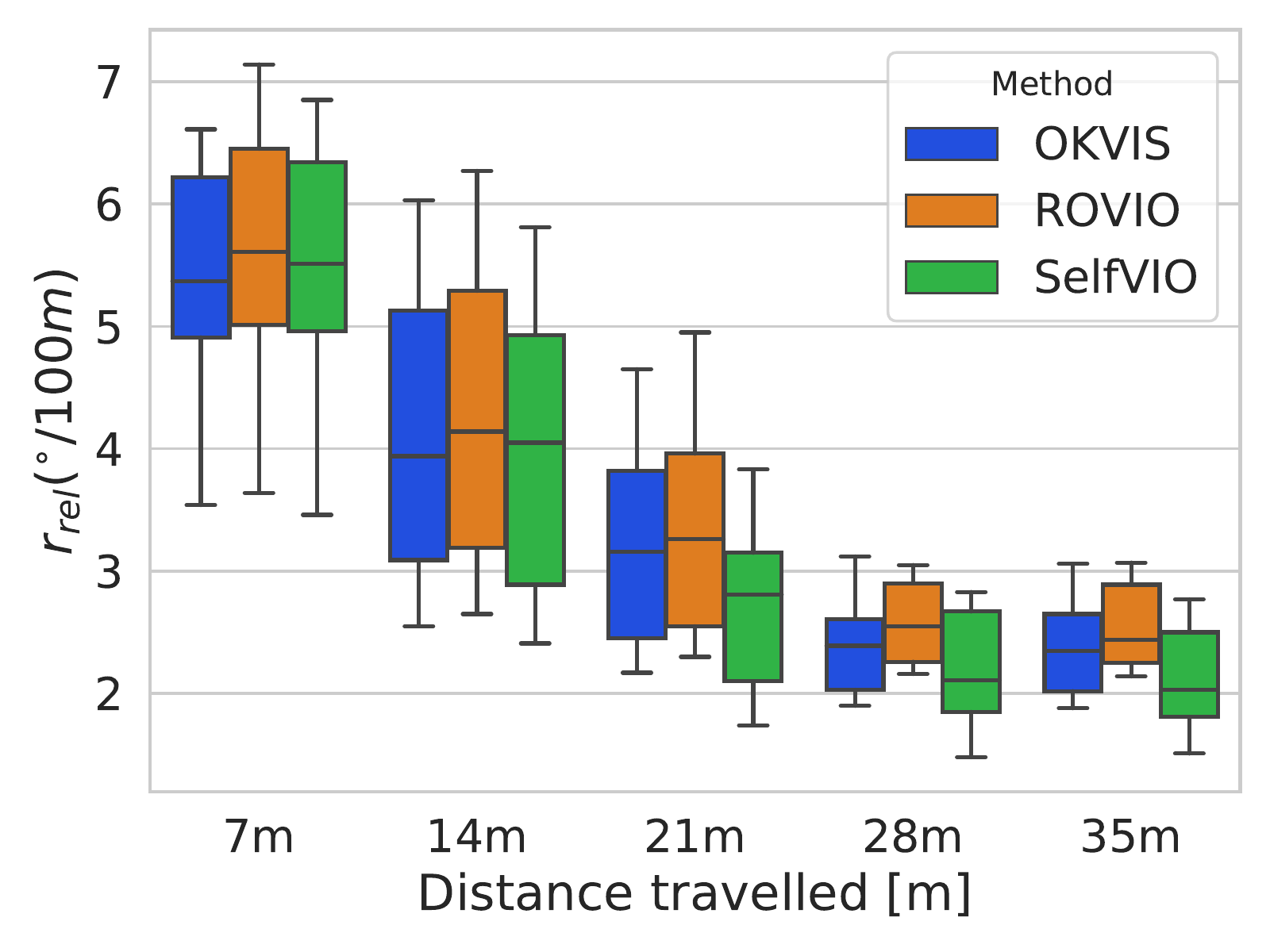}
        \caption{Rotational relative pose error} 
        \label{fig:euroc_rpe_rot}
	\end{subfigure}
	\caption{
{\color{black}
\textbf{Boxplot summarizing the relative pose error statistics with respect to the distance traveled for the monocular VIO pipelines on EuRoC dataset over all sequences.}  Errors are computed over trajectory segments of lengths $\{7, 14, 21, 28, 35\}$ m. We evaluate the monocular versions of the compared methods.
}}
    \label{fig:euroc_rpe}
\end{figure}

\subsection{Motion Estimation}
In this section, we comparatively discuss the motion estimation performance of the proposed method in terms of both vision-only and visual-inertial estimation modes.
\subsubsection{Visual Odometry} 
SelfVIO (no IMU) outperforms the VO approaches listed in Sec. \ref{subsec:evaluation} as seen in  Tab. \ref{tab:vo_traj},
which confirms that our results
are not due solely to our inclusion of IMU data.
{\color{black}
We evaluated monocular VISO2 and ORB-SLAM (without loop closure)
using full image resolution $1242\times376$ as they did not work with image resolution $416 \times 128$.
It should be noted that the results in Tab. \ref{tab:vo_traj} are for SelfVIO, VIOLearner, UnDeepVO, and SFMLearner networks that are tested on data on which they are also trained, which corresponds with the results presented in \cite{li2018undeepvo, shamwell2019unsupervised}. Although the sequences in Tab. \ref{tab:vo_traj} are used during the training, the results in Tab. \ref{tab:vo_traj} indicate the effectiveness of the supervisory signal as the unsupervised methods do not incorporate ground-truth pose and depth maps. We compare SelfVIO against UnDeepVO and VIOLearner using these results.
}

We also evaluate SelfVIO more conventionally by training on sequences $00-08$ and testing on sequences $09$ and $10$ that were not used in the training as was the case for \cite{zhou2017unsupervised, shamwell2019unsupervised}. These results are shown in Tab. \ref{tab:0910}. SelfVIO significantly outperforms SFMLearner on both KITTI sequences $09$ and $10$. 
{\color{black}
We also evaluate ORB-SLAM, OKVIS, and ROVIO using the full-resolution input images to show the effect of reducing the input size.
}

\subsubsection{Visual-Inertial Odometry}
The authors of VINet \cite{clark2017vinet} provide the errors in boxplots  compared to several state-of-the-art approaches for $100-500$ m on the KITTI odometry dataset. We reproduced the median, first quartile, and third quartile from \cite{clark2017vinet, shamwell2019unsupervised} to the best of our ability and included them in Tab. \ref{tab:vio_traj}. SelfVIO outperforms VIOLearner and VINet for longer trajectories ($100, 200, 300, 400, 500$m) on KITTI sequence 10.
Although SelfVIO (LSTM) is slightly outperformed by SelfVIO, it still performs better than VIOLearner and VINET, which shows CNN architecture in SelfVIO increases the estimation performance.
It should again be noted that
our network can implicitly learn camera-IMU extrinsic calibration from the data. 
We also compare SelfVIO against the traditional state-of-the-art VIO and include a custom EKF with VISO2 as in VINET \cite{clark2017vinet}. 

We successfully run SelfVIO on the KITTI odometry sequences 09 and 10 and include the results in Tab. \ref{tab:0910} and Fig. \ref{fig:traj_kitti_results}. 
SelfVIO outperforms OKVIS and ROVIO on KITTI sequences 09 and 10.
{\color{black}
However, both OKVIS and ROVIO require tight synchronization between the IMU measurements and the images that KITTI does not provide. This is most likely the reason for the poor performance of both approaches on KITTI. 
Additionally, the acceleration in the KITTI dataset is minimal, which causes a significant drift for the monocular versions of OKVIS and ROVIO.
These also highlight a strength of SelfVIO in that it can compensate for loosely temporally synchronized sensors without explicitly estimating their temporal offsets, showing the effectiveness of LSTM in the sensor fusion.
Furthermore, we evaluate ORB-SLAM, OKVIS, and ROVIO using the full-resolution images to show the impact of reducing the image resolution (see Tab. \ref{tab:0910}). Although higher resolution improves the odometry performance, OKVIS and ROVIO heavily suffer from loose synchronization, and ORB-SLAM is prone to large drifts without loop closure.
} 

In addition to evaluating with relative error over the entire
trajectory, we also evaluated SelfVIO RGB using RMSE over
five frame snippets as was done in \cite{zhou2017unsupervised, mahjourian2018unsupervised, shamwell2019unsupervised} for their similar
monocular approaches. 
As shown in Tab. \ref{tab:pose_snippet}, SelfVIO surpasses RMSE performance of SFMLearner, Mahjourian et al. and VIOLearner on KITTI trajectories 09 and 10.

The results on the EuRoC sequences are shown in Tab. \ref{tab:euroc_vio} and sample trajectory plots are shown in Fig. \ref{fig:traj_euroc_results}. 
SelfVIO produces the most accurate trajectories for
many of the sequences, even without explicit loop closing.
{\color{black}
We additionally evaluate the benchmark methods on the EuRoC dataset using full-resolution input images to show the impact of reducing the image resolution (see Tab. \ref{tab:euroc_vio}).
Unlike evaluation methods used in the supervised learning-based methods, we also evaluate SelfVIO on the sequences used for the training to show the effectiveness of the supervisory signal as SelfVIO does not incorporate ground-truth pose and depth maps. The test sequences are also shown with marks in Tab. \ref{tab:euroc_vio}, which are never used during the training.
In Fig. \ref{fig:euroc_rpe}, we show statistics for the relative translation and rotation error accumulated over trajectory segments of lengths $\{7, 14, 21, 28, 35\}$ m over all sequences for each platform-algorithm combination, which is well-suited for measuring the drift of an odometry system. These evaluation distances were chosen based on the length of the shortest trajectory in the EuRoC dataset, VR\_02 sequence with $36$ m.
}
To provide an objective
comparison to the existing related methods in the literature,
we use the following methods for evaluation described earlier in Section \ref{sec:rel_work}:
\begin{itemize}
\item MSCKF \cite{mourikis2007multi} - multistate constraint EKF,
\item SVO+MSF \cite{faessler2016autonomous} - a loosely coupled configuration of a visual odometry pose estimator \cite{forster2016svo} and an EKF for visual-inertial fusion \cite{lynen2013robust},
\item OKVIS \cite{leutenegger2015keyframe} - a keyframe optimization-based method using landmark reprojection errors,
\item ROVIO \cite{bloesch2017iterated} - an EKF with tracking
of both 3D landmarks and image patch features, and
\item VINS-Mono \cite{qin2018vins} - a nonlinear-optimization-based sliding window estimator using preintegrated IMU factors.
\end{itemize}
As we are interested in evaluating the odometry performance of the methods, no loop closure is performed. 
In difficult sequences (marked with D), the continuous inconsistency in brightness between the
images causes failures in feature matching for the filter based approaches, which can result in divergence of the filter.
On the easy sequences (marked with E), although OKVIS and VINSMONO slightly outperform the other methods, the accuracy of SVOMSF, ROVIO and SelfVIO
approaches is similar except that MSCKF has a larger error in the
machine hall datasets which may be caused by the larger scene
depth compared to the Vicon room datasets.

{\color{black}
As shown in Fig. \ref{fig:offset_euroc_results}, orientation offsets within a realistic range
of less than $10$ degrees show low numbers of errors and great
applicability of SelfVIO to sensor implementation with high degrees of miscalibration. 
}
Furthermore, offsets within
a range of less than $30$ degrees display a modestly sloped plateau that suggests successful learning of calibration.
In contrast, OKVIS shows surprising robustness to rotation errors under $20$ degrees but is unable to handle orientation
offsets around the $30$ degree mark, where 
error measures
appear to drastically increase. 
This is plausibly expected because deviations of this magnitude result
in large dimension shift, and unsurprisingly, OKVIS
appears unable to compensate.
{\color{black}
Furthermore, we evaluate SelfVIO, OKVIS, and VINS-Mono on miscalibrated data subject to various translational and temporal offsets between visual and inertial sensors.
Table \ref{tab:euroc_calib} shows that VINS-Mono and OKVIS perform poorly as the translation and time offsets increase, which is due to their need for tight synchronization. 
SelfVIO achieves the smallest relative translational and rotational errors under various temporal and translational offsets, which indicates the robustness of SelfVIO against loose temporal and spatial calibration.
OKVIS fails to track in sequences MH02-05 when the time offset is set to be $90$ ms. We have also tested larger time offsets such as $120$ ms, but neither OKVIS nor VINS-Mono provides reasonable estimates.
}

{\color{black}
By explicitly modeling the sensor fusion process, we demonstrate the strong correlation between the
odometry features and motion dynamics. 
Figure \ref{fig:attention-vis} illustrates that features extracted
from visual and inertial measurements are complementary in various conditions. The contribution of inertial features increases in the presence of fast rotation. In contrast, visual features are highly active during large translations, which provides insight
into the underlying strengths of each sensor modality.
}

\section{Conclusion}
\label{sec:conclusion}

In this work, we presented our SelfVIO architecture
and demonstrated superior performance against state-of-the-art VO, VIO,
and even VSLAM approaches. 
Despite using only monocular source-target image pairs, SelfVIO
surpasses state-of-the-art depth and motion estimation performances  
of both traditional and learning-based approaches such as VO, VIO and VSLAM that use sequences
of images, keyframe based bundle adjustment, and full bundle
adjustment and loop closure. This is enabled by a
novel adversarial training and visual-inertial sensor fusion technique embedded in our end-to-end trainable deep
visual-inertial architecture.
Even when IMU data are not
provided, SelfVIO with RGB data outperforms deep
monocular approaches in the same domain.
In future work, we plan to develop a stereo version of SelfVIO that could utilize the disparity map.


\ifCLASSOPTIONcaptionsoff
  \newpage
\fi



\bibliographystyle{IEEEtran}
\bibliography{IEEEabrv,mybibfile}
%
%
%

%

\begin{IEEEbiographynophoto}{Yasin Almalioglu}
Yasin Almalioglu is currently enrolled as a DPhil (PhD) student in computer science at the University of Oxford since January 2018.
Yasin Almalioglu received the BSc degree with honours in computer engineering from Bogazici University, Istanbul, Turkey in 2015. He was research intern at CERN Geneva, Switzerland and Astroparticle and Neutrino Physics Group at ETH Zurich, Switzerland in 2013 and 2014, respectively. He was awarded the Engin Arik Fellowship in 2013. He received the MSc degree with high honours in computer engineering from Bogazici University, Istanbul, Turkey in 2017. 
\end{IEEEbiographynophoto}

\begin{IEEEbiographynophoto}{Mehmet Turan}
Mehmet Turan received his diploma degree from RWTH Aachen University,
Germany in 2012 and his PhD degree from ETH Zurich, Switzerland in 2018.
Between 2013-2014, he was a research scientist at UCLA (University of
California Los Angeles). Between, 2014-2018 he was a research scientist
at Max Planck Institute for Intelligent Systems and between 2018-2019,
he was a postdoctoral researcher at the Max Planck Institute for
Intelligent Systems. He received the DAAD (German Academic Exchange
Service) fellowship between years 2005–2011 and Max Planck fellowship
between 2014–2019. He has also received Max Planck-ETH Center for
Learning Systems fellowship between 2016–2019. Currently, he is a faculty at the
Institute of Biomedical Engineering, Bogazici University, Turkey. 
\end{IEEEbiographynophoto}

\begin{IEEEbiographynophoto}{Alp Eren Sar{\i}}
Alp Eren Sar{\i} is a M.Sc. student in electrical and electronics engineer, Middle East Technical University. His research focuses on computer vision and signal processing.
He obtained the B.Sc. degree from electrical and electronics engineer, Middle East Technical University in 2018.
\end{IEEEbiographynophoto}

\begin{IEEEbiographynophoto}{Muhamad Risqi U. Saputra}
Muhamad Risqi U. Saputra is a DPhil student at Department of Computer Science, University of Oxford, working on vision-based indoor positioning system.
He previously received bachelor and master degree from Department of Electrical Engineering and Information Technology, Universitas Gadjah Mada, Indonesia, in 2012 and 2014 respectively. His research interest revolves around computer vision and machine learning applied to indoor positioning and mapping, healthcare, assistive technology, and education.
\end{IEEEbiographynophoto}

\begin{IEEEbiographynophoto}{Pedro P. B. de Gusmão}
Pedro P. B. de Gusmão currently holds a postdoctoral research position at the Cyber-Physical Systems group, working on the NIST grant “Pervasive, Accurate, and Reliable Location Based Services for Emergency Responders".
He received my bachelor’s degree in Telecommunication Engineering from the University of São Paulo and my master’s degree on the same field from the Politecnico di Torino in a double degree program. During MSc studies, he was a recipient of an Alta Scuola Politecnica scholarship; and in 2017, he obtained my PhD from the same Politecnico.
His research interests include computer vision, machine learning and signal processing.  
\end{IEEEbiographynophoto}

\begin{IEEEbiographynophoto}{Andrew Markham}
Andrew Markham is an Associate Professor and he works on sensing systems, with applications from wildlife tracking to indoor robotics to checking that bridges are safe. He works in the cyberphysical systems group. He designs novel sensors, investigate new algorithms (increasingly deep and reinforcement learning-based) and applies these innovations to solving new problems. Previously he was an EPSRC Postdoctoral Research Fellow, working on the UnderTracker project. He obtained his PhD from the University of Cape Town, South Africa in 2008 researching the design and implementation of a wildlife tracking system, using heterogeneous wireless sensor networks.
\end{IEEEbiographynophoto}

\begin{IEEEbiographynophoto}{Niki Trigoni}
Niki Trigoni is a Professor at the Oxford University Department of Computer Science and a fellow of Kellogg College. She obtained her DPhil at the University of Cambridge (2001), became a postdoctoral researcher at Cornell University (2002-2004), and a Lecturer at Birkbeck College (2004-2007). At Oxford, she is currently Director of the EPSRC Centre for Doctoral Training on Autonomous Intelligent Machines and Systems, a program that combines machine learning, robotics, sensor systems and verification/control. She also leads the Cyber Physical Systems Group, which is focusing on intelligent and autonomous sensor systems with applications in positioning, healthcare, environmental monitoring and smart cities.
\end{IEEEbiographynophoto}




\end{document}